\documentclass[journal]{IEEEtran}
\usepackage{times}
\usepackage{epsfig}
\usepackage{graphicx}
\usepackage{amsmath}
\usepackage{amssymb}
\usepackage{graphicx}
\usepackage{amssymb}
\usepackage{booktabs}
\usepackage{graphicx}
\usepackage{float}
\usepackage{stfloats}
\usepackage{epstopdf}
\usepackage{algorithm,algorithmic}
\usepackage{multirow}
\usepackage{bbm}
\usepackage{framed}
\usepackage{amsthm}
\usepackage{adjustbox}
\usepackage{color}
\usepackage{lineno}
\usepackage{url}
\usepackage{cleveref}
\usepackage{cite}
\usepackage{colortbl}
\usepackage{booktabs}
\usepackage{subfigure}
\usepackage{color}
\makeatletter
\newif\if@restonecol
\makeatother

\usepackage{enumitem}

\begin{document}
%
\title{Universal Domain Adaptation for Remote Sensing Image Scene Classification}
%
%
%

\author{Qingsong Xu,
         Yilei Shi,~\IEEEmembership{Member,~IEEE}, Xin Yuan,~\IEEEmembership{Senior Member,~IEEE},
        and~Xiao Xiang Zhu,~\IEEEmembership{Fellow,~IEEE}

\thanks{This work is jointly supported by the German Research Foundation (DFG GZ: ZH 498/18-1; Project number: 519016653), by the European Research Council (ERC) under the European Union's Horizon 2020 research and innovation programme (grant agreement No. [ERC-2016-StG-714087], Acronym: \textit{So2Sat}), by the Helmholtz Association under the Framework of the Helmholtz Excellent Professorship ``Data Science in Earth Observation - Big Data Fusion for Urban Research''(grant number: W2-W3-100), by the German Federal Ministry of Education and Research (BMBF) in the framework of the international future AI lab "AI4EO -- Artificial Intelligence for Earth Observation: Reasoning, Uncertainties, Ethics and Beyond" (grant number: 01DD20001) and by German Federal Ministry for Economic Affairs and Climate Action in the framework of the "national center of excellence ML4Earth" (grant number: 50EE2201C). The work of X. Yuan is supported by the National Natural Science Foundation of China (grant number: 62271414), Zhejiang Provincial Natural Science Foundation of China (grant number: LR23F010001), Westlake Foundation (grant number: 2021B1 501-2) and the Research Center for Industries of the Future (RCIF) at Westlake University. (Qingsong Xu
and Yilei Shi contributed equally.) (Corresponding
author: Xiao Xiang Zhu.)}
\thanks{Q. Xu, and X. X. Zhu are with the Chair of Data Science in Earth Observation, Technical University of Munich (TUM), 80333 Munich, Germany. (e-mails: qingsong.xu@tum.de; xiaoxiang.zhu@tum.de).}
\thanks{Y. Shi is with the Chair of Remote Sensing Technology, Technical University of Munich (TUM), 80333 Munich, Germany. (e-mail: yilei.shi@tum.de).}
\thanks{X. Yuan is with the School of Engineering, Westlake University, Hangzhou,
	Zhejiang 310030, China (e-mail: xyuan@westlake.edu.cn).}
 }

\markboth{IEEE TRANSACTIONS ON GEOSCIENCE AND REMOTE SENSING}%
{Shell \MakeLowercase{\textit{et al.}}: Bare Demo of IEEEtran.cls for Journals}

\maketitle

\begin{abstract}
\textcolor{blue}{This work has been accepted by IEEE TGRS for
publication.} 
The domain adaptation (DA) approaches available to date are usually not well suited for practical DA scenarios of remote sensing image classification, since these
methods (such as unsupervised DA) rely on rich prior knowledge about the relationship between label sets of source and target domains, and source data are often not accessible  due to privacy or confidentiality issues.
To this end, we propose a practical universal domain adaptation setting for remote sensing image scene classification that requires no prior knowledge on the label sets. Furthermore,  a novel universal domain adaptation method without source data is proposed for cases when the source data is unavailable. The architecture of the model is divided into two parts: the source data generation stage and the model adaptation stage. The first stage estimates the conditional distribution of source data from the pre-trained model using the knowledge of class-separability in the source domain and then synthesizes the source data. With this synthetic source data in hand, it becomes a universal DA task to classify a target sample correctly if it belongs to any category in the source label set, or mark it as ``unknown" otherwise. In the second stage, a novel transferable weight that distinguishes the shared and private label sets in each domain promotes the adaptation in the automatically discovered shared label set and recognizes the ``unknown'' samples successfully. Empirical results show that the proposed model is effective and practical for remote sensing image scene classification, regardless of whether the source data is available or not. \textcolor{red}{The code is available at \url{https://github.com/zhu-xlab/UniDA}.}

\end{abstract}


%
\IEEEpeerreviewmaketitle

\section*{Nomenclature}
\addcontentsline{toc}{section}{Nomenclature}
\begin{IEEEdescription}[\IEEEusemathlabelsep\IEEEsetlabelwidth{$V_1,V_2,V_3$}]
\item[$M$] Pre-trained model on real source domain.
\item[$F$] Feature extractor. $\theta_{f}$ denotes parameters of $F$.
\item[$C$] Classifier. $\theta_{c}$ denotes parameters of $C$.
\item[$D^{\prime}$] Non-adversarial domain discriminator. $\theta_{d^{\prime}}$ denotes parameters of $D^{\prime}$.
\item[$D$] Adversarial domain discriminator. $\theta_{d}$ denotes parameters of $D$.
\item[$p(x)$] Real source domain distribution.
\item[$q(x)$] Target domain distribution.
\item[$p(y_{s})$] True labeled distribution of the source domain.
\item[$y$] One-hot vector that represents a label.
\item[$p_{y}(y)$] Estimated categorical distribution of the true labeled distribution $p(y_{s})$.
\item[$z$] Standard normal vector of a low-dimensional noise.
\item[$p_{z}(z)$] Multivariate Gaussian distribution describing the source data points.
\item[$p(x\mid y,z)$] Conditional distribution of source data.
\item[$G$] Generator that obtains the empirical distribution $p(x\mid y, z)$. $\theta_{g}$ denotes parameters of $G$.
\item[$x_{s}$] Source image.
\item[$x_{f}$] Synthetic source image.
\item[$x_{t}$] Target image.
\item[$D_{s}$] Real source domain.
\item[$D_{f}$] Synthetic source domain. $|D_{f}|$ denotes the size of synthetic source domain.
\item[$D_{t}$] Target domain.
\item[$Y_{f}$] Label set of the synthetic source domain.
\item[$Y_{t}$] Label set of the target domain.
\item[$\overline{Y_{f}}$] Private label set of the synthetic source domain.
\item[$\overline{Y_{t}}$] Private label set of the target domain.
\item[$\xi$] Jaccard index.
\item[$\bar{y}$] Probability vector predicted by the classifier $C$.
\item[$\phi_{j}(x)$] Activation at the $j$th layer of the style loss network.
\item[$G_{j}^{\phi}(x)$] Gram matrix.
\item[$w_{f}(x)$] Sample-level  transferable weight for synthetic  source data (scalar).
\item[$w_{t}(x)$] Sample-level  transferable weight for target data (scalar).
\item[$w_{0}$] Decision threshold (scalar).
\item[$d(x)$] Domain similarity of  target domain samples to the synthetic source domain samples (scalar).
\item[$\max \bar{y}(x)$] Confidence of predicted probabilities (scalar).
\item[$\ell_{\mathrm{cls}}$] Classifier loss.
\item[$\ell_{\mathrm{style}}$] Style loss.
\item[$\ell_{\mathrm{adv}}$] Adversarial loss function for adaptation.
\item[$\ell_{\mathrm{ce}}^{f}$] Cross-entropy loss on the synthetic source domain.
\item[$\ell_{\mathrm{simi}}$] Binary cross-entropy loss for non-adversarial domain discriminator.
\end{IEEEdescription}

\section{Introduction}
%
%
%
%
\label{sec:introduction}
\IEEEPARstart{R}{emote} sensing image scene classification is a procedure for assigning semantic labels according to the content of remote sensing scenes~\cite{tuia2021recent}, which is beneficial to  traffic analysis, urban area monitoring and planning~\cite{liu2017classifying, zhu2022knowledge}, land-use and land-cover~\cite{zhu2022land}, and
hazard detection and avoidance~\cite{xu2022mffenet}, among other applications. In recent years, many deep learning approaches have been proposed for scene classification of remote sensing images~\cite{cheng2020remote,9440852}, such as autoencoder~\cite{yao2016semantic}, convolutional neural networks (CNNs)~\cite{xie2019scale}, generative adversarial networks (GANs)~\cite{yu2019attention}, prototype-based memory networks~\cite{hua2021}, and transformer~\cite{tan2022transformer}. These methods usually assume that the training and testing data share the same distribution. However, in a real application, due to the influence of sensors, geographic locations, imaging conditions, and other factors, the distribution of training and testing data may be different. 
This phenomenon is referred to as the domain gap~\cite{ben2010theory}. 
To  address the domain gap problem among different datasets, domain adaptation (DA) algorithms have been proposed.
DA aims to leverage a source domain to learn a model that performs well on
a different but related  target domain~\cite{saenko2010adapting}. 

\begin{figure*}[t]
	\centering
	{\includegraphics[width = .82\textwidth]{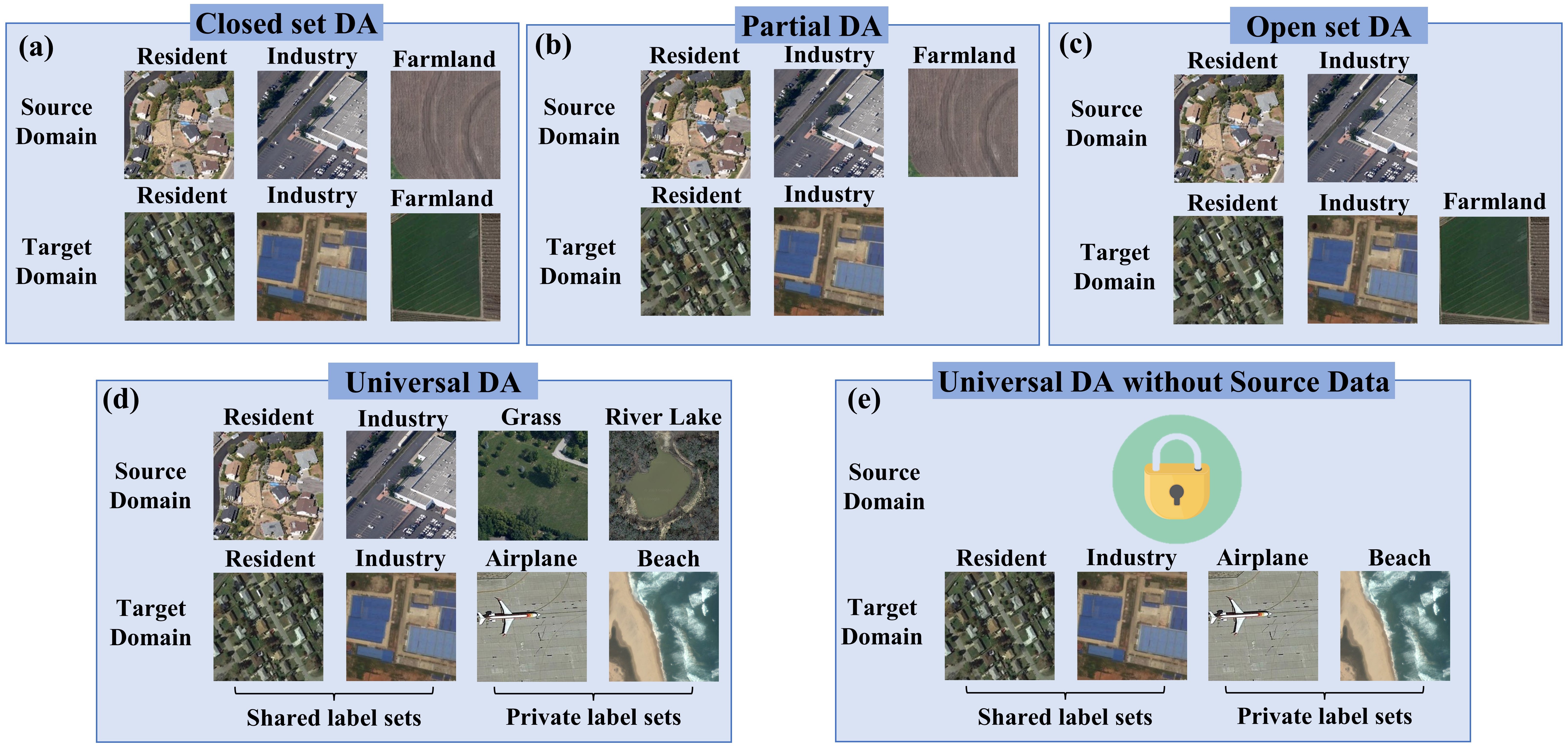}}
	\caption{Different domain adaptation scenarios. (a) Closed-set DA, which assumes that the source domain and the target domain have shared label sets. (b) Partial DA, which assumes that  target label sets are considered a subset of source label sets. (c) Open-set DA, which assumes that source label sets are considered a subset of target label sets. (d) Universal DA, which imposes no prior knowledge on the label sets.  Label sets are divided into shared and private label sets in each domain. (e) Universal DA without source data. The source dataset is not available in the practical universal DA scenarios of remote sensing.}
	\label{fig:1}
\end{figure*}

In remote sensing scene classification, most  existing DA approaches~\cite{xu2020class, wittich2021appearance} are proposed to tackle the domain gap between different domains by learning a domain invariant feature representation. Based on the knowledge of the relationship between the source and target label space (category-gap), DA can be divided into closed-set DA, partial DA, and open-set
DA. Specifically, closed-set DA usually addresses the domain adaptation problem by leveraging the adversarial learning behaviors of GANs to perform distribution alignment in the
pixel, feature, and output spaces~\cite{tasar2020standardgan, xu2020class, guo2021unsupervised, xu2022mffenet}, which assumes a
shared label set between the source and target domains, as shown in Fig. \ref{fig:1}(a).
In order to relax this assumption, two alternatives have been proposed: partial DA~\cite{cao2018partial}, in which the target label space is considered a subset of the source label space, as shown in Fig.~\ref{fig:1}(b), and open-set
DA~\cite{saito2018open}, in which the source label space is considered a subset of the target label space, as shown in Fig.~\ref{fig:1}(c). For example, an open-set domain adaptation algorithm~\cite{zhang2021open} is proposed in which transferability and discriminability are explored for the purpose of remote sensing image scene classification. However, these DA methods have two major bottlenecks in the domain adaptation of remote sensing scene classification in the wild.

\begin{itemize}[leftmargin= 10 pt, itemsep= 0 pt, topsep = 2 pt, parsep= 2 pt]
	\item  In a general scenario, we cannot select the proper domain adaptation methods (closed-set DA, partial DA, or open-set
	DA) because no prior knowledge about the target domain label set is given.
	\item  The source dataset is not available in many practical application scenarios of remote sensing. For example, many satellite companies and users will only provide pre-trained models instead of their source data due to data privacy and security issues. In addition, the source datasets, like high-resolution remote sensing images, may be so large that it is not practical or convenient to transfer or retain them to different platforms.
\end{itemize}

\begin{figure*}[t]
	\centering
	{\includegraphics[width = .8\textwidth]{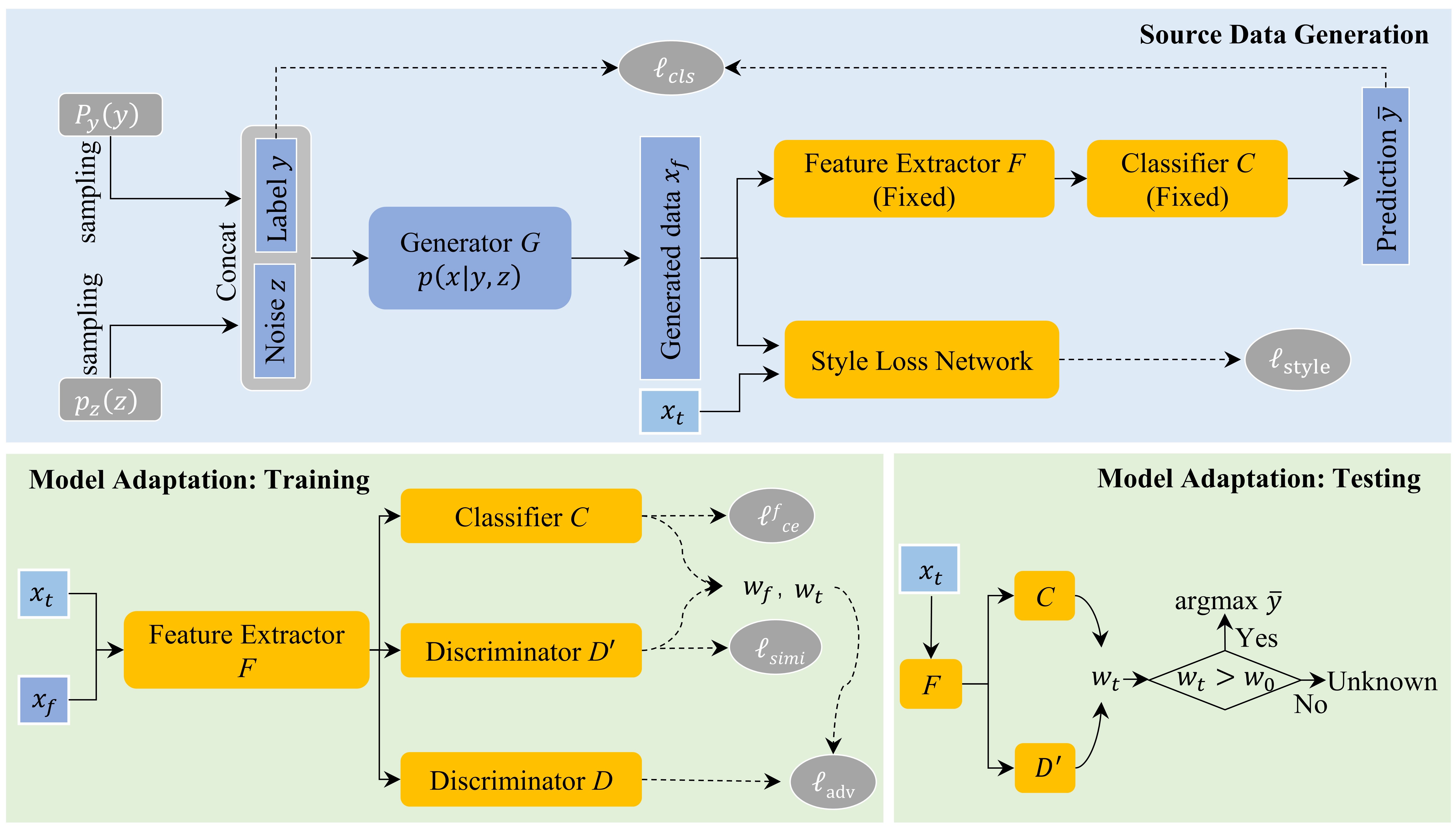}}
	\vspace{-4mm}
	\caption{Overview of the proposed UniDA without source data (SDG-MA). The model consists of a source data generation stage and a model adaptation stage.}
	\label{fig:2}
\end{figure*}
To address the first challenge, a novel scenario of universal domain adaptation (UniDA) is proposed. As shown in Fig.~\ref{fig:1}(d), UniDA removes all constraints and includes all the above adaptation settings~\cite{you2019universal}. UniDA may contain a shared label set and hold a private label set for a given source label set and a target label set.
 Two challenges are exposed in a UniDA setting. (1) If we naively match the entire source domain with the entire target domain, the mismatch of different label sets will deteriorate the model. Thus, the samples coming from the shared label set between the source and target domains should be automatically detected and matched. (2) The target samples from private label sets should be marked as ``unknown" since there are no labeled training data for these classes. Currently, different transferability criteria (such as entropy in ~\cite{you2019universal}, pseudo-margin vector in~\cite{yin2021pseudo}, and the mixture of entropy, confidence, and consistency in~\cite{fu2020learning}) have been proposed to distinguish samples from  shared label sets and those in private label sets in the field of computer vision.
To address the second challenge, in computer vision,  source-free domain adaptation is under continuous exploration~\cite{kundu2020towards,liang2020we, ding2022source,xu2022universal}. For example, ~\cite{kundu2020universal} proposes the universal source-free domain adaptation setting for natural image classification.
 However, existing UniDA methods~\cite{you2019universal,kundu2020universal,yin2021pseudo,fu2020learning} in computer vision normally assume that
the source data set is available when building the classifier platform. This assumption is not  valid and practical for  the second challenge. Thus, developing a universal domain adaptation method without source data (Fig.~\ref{fig:1}(e)) has a practical value and is thus desired in real application scenarios of remote sensing image classification. 

In UniDA without source data, pre-trained models can be available. Pre-trained models not only serve as strong baselines for the original dataset, but also contain knowledge of the original dataset. Therefore, generating {\em synthetic source domain data} from the pre-trained model is the first problem to be solved.  There are recent works for distilling a network’s knowledge by a small dataset~\cite{li2018knowledge} or no observable data~\cite{yoo2019knowledge}. It is worth noting that we cannot use generative adversarial networks to directly generate artificial data (similar to~\cite{li2020model}), because the core of UniDA without source data is to restore the category distribution (including the shared label set and the private label set) from the pre-trained model.

Bearing these concerns in mind, we propose the {\bf UniDA without source data} in order to introduce the UniDA setting into remote sensing datasets. In this case, we merely have access to the pre-trained model from the source domain. We have no information about the source data distribution that was used to train. UniDA without source data poses two major technical challenges for designing the corresponding models in the wild. (1) Distilling the  knowledge of source data from the pre-trained model. The knowledge is consistent with the source in the category distribution (including the shared label set and the private label set), and is as close as possible to the target in style. (2) Domain adaptation should be applied to align distributions of the synthetic source and target data in the technical challenges of UniDA.

To address these two challenges, our proposed UniDA without source data for remote sensing images consists of a source data generation (SDG) stage and a model adaptation (MA) stage. In the SDG stage, we reformulate the goal as estimating the conditional distribution rather than the distribution  of the source data, since the source data space is exponential with the dimensionality of data. After the conditional distribution of the source data is obtained, a well-defined criterion can be used to distinguish different degrees of uncertainty in order to separate the target samples from the shared label set and those from the private label. However, uncertainty is usually measured by entropy~\cite{you2019universal, saito2020universal}, which lacks discriminability for uncertainty when the categorical distributions are relatively uniform~\cite{fu2020learning}. Thus, a novel {\em transferable weight} is defined by considering confidence and domain similarity.

In a nutshell, our contributions are as follows:
\begin{itemize}[leftmargin= 10 pt, itemsep= 0 pt, topsep = 2 pt, parsep= 2 pt]
	\item We introduce a more practical and challenging UniDA setting for remote sensing image scene classification.
	\item We propose a new UniDA model (SDG-MA), which is composed of a source data generation stage and a model adaptation stage.
	\item In order to generate reliable source domain samples, a  novel {\em conditional probability recovery method} of the source domain is designed to distill  category knowledge.
	\item A novel {\em transferable weight} is utilized to distinguish the shared label sets and the private label sets in each domain.
	\item Experimental results on four UniDA settings for remote sensing image scene classification demonstrate that the proposed model is effective and practical, regardless of whether the source domain is available or not. 
\end{itemize}
\section{Related Work}
Most existing DA settings for remote sensing image scene classification can be summarized as closed-set, partial, and open-set DA based on the label set relationship.
Closed-set DA is a scenario where the source and target domains share the same label set. The main challenge in this scenario is to overcome the domain gap that comes as a result of the samples being taken from different distributions. 
Among the recent work on  closed-set DA for remote sensing, adversarial learning frameworks have attracted significant interest because of the improved quality of alignment between distributions by adapting representations of different domains. 
GANs are commonly used at feature maps generated from CNNs where a domain discriminator is trained to correctly classify the domain of each input feature. For example, domain-adversarial neural networks (DANN)~\cite{7954001}, Siamese GAN~\cite{rs10020351}, Attention GAN~\cite{yu2019attention}, and domain adaptation via a task-specific classifier (DATSNET) framework~\cite{9714321} are presented for the classification of remote sensing images, by  learning an invariant  representation.  Recently, a multitude of  closed-set DA algorithms for  remote sensing image scene classification~\cite{7917346,8291513,8794530,8651371,zhang2020domain,guo2022semi,HUANG2023192} is designed to reduce the global or local distribution differences between domains. In addition, closed-set DA with multiple source domains~\cite{lu2019multisource} is proposed for remote sensing image classification. However, it is difficult to ensure that the source domain and the target domain have common classes. Thus, partial DA and open-set DA are proposed to relax this limitation.  Partial DA handles
the case where the target classes are a subset of source classes. This task is solved by performing importance-weighting on source examples that are similar to samples in the target~\cite{cao2018partial,cao2019learning,liang2020balanced}. Open set
DA is a more realistic version, where the new classes will appear in the target domain. In the  open set
DA setting, the target domain contains unknown classes that do not present in the source domain. In remote sensing image scene classification, an open set DA algorithm via exploring transferability and discriminability (OSDA-ETD)~\cite{zhang2021open} is  proposed to reduce the distribution discrepancy of the same classes in different domains and enlarge the distribution discrepancy of different classes in different domains. In addition, some open set DA networks based on adversarial learning~\cite{9323944,adayel2020deep,nirmal2020open,tang2021open} and graph convolutional networks~\cite{9953105,zhao2022graph} are presented for remote sensing image scene classification. However, almost all these methods rely on prior knowledge about the relationship between label sets of source and target domains and assume the co-existence of source and target data.  Thus, in order to promote the development of DA methods, we propose a general setting (UniDA) for remote sensing image scene classification.

\section{Methodology}
In this section, we elaborate the problem of the UniDA setting without source data and address it by a novel dual-stage framework (SDG-MA), shown in Fig.~\ref{fig:2}. 
\subsection{Problem Setting}
For UniDA setting without source data (SDG-MA), we merely have access to the pre-trained model $M$, including feature extractor $F$ and classifier $C$. We have no information about the source data distribution $p(x)$ that is used to train $M$. Thus, considering the MA in the second stage, our first goal is to generate reliable source data $x_{f}$ from the pre-trained model $M$. The synthetic distribution is consistent with the source data distribution $p(x)$ in the category distribution (including the shared label set and the private label set), and is as close as possible to the target domain in style.
However, it is impracticable to estimate $p(x)$ directly since the source data space is exponential with the dimensionality of data. 
Thus, as shown in  the source data generation stage of Fig.~\ref{fig:2},  we generate the set by modeling a conditional probability of $x$ given two random vectors $y$ and $z$. $y$ ($y \sim p_{y}(y)$) is a probability vector that represents a label, where $p_{y}(y)$ is an estimation of the true labeled distribution $p(y_{s})$ of the source domain. $z$ ($z \sim p_{z}(z)$) is a low-dimensional noise, where  $p_{z}(z)$ is a random distribution describing the source data points. Thus, we reformulate the goal as to estimate the conditional distribution of source data $p(x\mid y,z)$ instead
of the distribution $p(x)$. 

After obtaining the conditional distribution of source data $p(x\mid y,z)$  from SDG stage, it becomes a UniDA task but now with synthetic source domain. Our second goal is to align distributions of the synthetic source domain and target domain in the technical challenges of domain gap and category gap.
A synthetic source domain and a target domain are represented by $D_{f} = \left\{\left(x_{f}^{i}, y_{f}^{i}\right) \sim p(x\mid y,z) \right\}_{i=1}^{n_{f}}$ sampled from conditional distribution $p(x \mid y, z)$ and  $D_{t} = \left\{\left(x_{t}^{i}\right) \sim q(x)\right\}_{i=1}^{n_{t}}$ sampled from target distribution $q(x)$, respectively. We denote by $Y_{f}$ ($Y_{t}$) the label
set of the synthetic source (target) domain. The shared label set is denoted by $Y=Y_{f} \cap Y_{t}$. The private label sets of the source and target domain are represented by $\overline{Y_{f}}=Y_{f} \backslash Y$ and $\overline{Y_{t}}=Y_{t} \backslash Y$, respectively. The Jaccard index of the label sets of the two domains, $\xi=\frac{|Y|}{|Y_{f} \cup Y_{t}|}$,  is used to measure the overlap in classes.

For UniDA setting with source data (MA), the real source domain $D_{s} = \left\{\left(x_{s}^{i}, y_{s}^{i}\right) \sim p(x) \right\}_{i=1}^{n_{s}}$ is available. Thus, only the MA stage is used to align distributions of the real source domain and target domain in the technical challenges of UniDA.
\subsection{Source Data Generation}
Source data generation includes two modules, conditional probability generation module and data diversity module. Specifically, firstly, conditional probability generation module is presented to prove that the conditional distribution of source data $p(x \mid y, z)$ can be estimated by estimating the categorical likelihood $p(y\mid x)$ and the property likelihood $p(z \mid x)$. Secondly, in order to generate a reliable source domain for UniDA, the generated data $x_{f}$ must meet two conditions: 1) in data content, all category distributions in the pre-trained model $M$ can be restored, including source-share and source-private category distributions, and 2) in data style, the generated data can remain similar to the target domain style distribution. Thus, to meet these two conditions,  a data diversity module is proposed to ensure the data diversity of the generated source domain. In addition, different schemes of data diversity generation are
compared in Section~\ref{Sec:model}.
\subsubsection{Conditional Probability Generation Module}
Recall that $y$ ($y \sim p_{y}(y)$) and $z$ ($z \sim p_{z}(z)$) are a probability vector of a source distribution and a low-dimensional noise, respectively. The variables $y$ and $z$ are conditionally independent of each other given source data $x$, since they both depend on $x$ but have no direct interactions. 
In order to generate a reliable and balanced source domain $D_{f}$,  the probability of each sampled point $x$ is $1 / |D_{f}|$, and the probability at any other point is zero. Thus, $D_{f}=\{{\arg \max}_x p(x \mid y, z)\}$.
Based on Bayesian theory~\cite{bernardo2009bayesian,yoo2019knowledge}, the ${\arg \max}_x p(x \mid y, z)$ can be expressed as follows:
\begin{equation}
\begin{aligned}
&\arg \max_{x} p(x \mid y, z) \\
&=\arg \max_{x} (\log p(y \mid x, z)+\log p(x \mid z) -\log p(y \mid z)) \\
&=\arg \max_{x} (\log p(y \mid x) + \log p(y \mid z) \\
& \quad+\log p(x \mid z)-\log p(y \mid z)) \\
&=\arg \max _{x}(\log p(y \mid x)+\log p(x \mid z)) \\
&=\arg \max _{x}(\log p(y \mid x)+\log p(z \mid x)\\
&\quad +\log p(x)-\log p(z)) \\
& \approx \arg \max _{x}(\log p(y \mid x)+\log p(z \mid x)).
\end{aligned}
\end{equation}
In this way, the distribution $p(x \mid y, z)$ can be estimated by estimating the categorical likelihood $p(y\mid x)$ of the variable $y$ given $x$ and the property likelihood $p(z \mid x)$ of the variable $z$ given $x$. Thus, as shown in the `Source Data Generation' module in Fig.~\ref{fig:2}, a generator $G$ is designed to obtain the empirical distribution $p(x\mid y, z)$ by combining $y$ and $z$ randomly sampling from the distributions $p_{y}(y)$ and $p_{z}(z)$. In our experiments, we set $p_{y}(y)$ to the random categorical distribution of source domain that produces one-hot vectors as $y$, and $p_{z}(z)$ to the multivariate
Gaussian distribution that produces standard normal vectors as $z$.
\subsubsection{Data Diversity Module}
First, in order to recover the data content from the pre-trained model $M$, a  classifier loss $\ell_{\mathrm{cls}}$ is designed. Specifically, given a sampled class vector $y$ and a sampled noise vector $z$ as inputs, $G$ is trained to produce a synthetic source domain sample  that $M$ is likely to classify as $\bar{y}$. The classifier loss can force the generated data to follow the similar class distribution from model $M$, by minimizing the distance between $y$ and  $\bar{y}$, which  can be formulated as follows:
\begin{equation}
\ell_{\mathrm{cls}}(y, \bar{y})=-\sum_{i \in Y_{f}} y_{i} \log M(G(y, z))_{i}.
\end{equation}
Notably, $y$ and $\bar{y}$ are not scalars but probability vectors of length $Y_{f}$. Thus, the {\em cross-entropy} between two probability distributions is utilized to measure the distance between $y$ and $\bar{y}$.

However, the classifier loss $\ell_{\mathrm{cls}}$ easily leads to generate similar data points for each class in the synthetic source domain. Furthermore, it is necessary for domain adaptation to  transfer synthetic source images to the target style. A  style loss $\ell_{\mathrm{style}}$ is presented to measure differences in style
between a synthetic source image $x_{f}$ and a target image $x_{t}$. The style of remote sensing images represents colors, textures,
edges, common patterns, and other image style descriptions. Concretely, we make use of a 16-layer VGG network pre-trained on the ImageNet~\cite{simonyan2014very}  to  measure multi-scale feature style differences between images, which can be described as:
\begin{align}
&\ell_{\mathrm{style }}(x_{f}, x_{t})=\sum_{j=1}^{4}\left\|G_{j}^{\phi}(x_{f})-G_{j}^{\phi}(x_{t})\right\|_{F}^{2},\\
&G_{j}^{\phi}(x)=\frac{1}{C_{j} H_{j} W_{j}} \sum_{h=1}^{H_{j}} \sum_{w=1}^{W_{j}} \phi_{j}(x)_{c, h, w} \phi_{j}^{T}(x)_{c, h, w},
\end{align}
where $\phi_{j}(x)$ is the activation at the $j$th layer of the style loss network, and is a feature map of shape $C_{j} \times H_{j} \times W_{j}$. $G_{j}^{\phi}(x)$ denotes a Gram matrix that is equal to the average value of the product of the feature and the transposition of the feature.  The Gram matrix can grasp the general style of the entire image. The style loss $\ell_{\mathrm{style }}(x_{f}, x_{t})$ is  the squared Frobenius norm of the difference between the Gram matrices of  synthetic source image $x_{f}$ and target image $x_{t}$. In addition, different layers have different feature styles in the VGG network. Therefore, we sum the Gram matrices difference for each of the four activation layers in the VGG-16. 

\subsection{Model Adaptation}
The objective of MA is to update the pre-trained model $M$, which distinguishes samples from the target shared label set $Y$ and those in the target private label set $\overline{Y_{t}}$. One important challenge for  UniDA is detecting transferable samples. In order to address this challenge, the sample transferable weight $w_{f}(x_{f})$ or $w_{t}(x_{t})$ is utilized during the training stage to estimate the confidence that $x_{f}$ or $x_{t}$ is from the shared label set. 
Furthermore, during the testing stage, we use the transferable weight as a decision threshold $w_{0}$ to decide whether we should predict a class or  mark the sample as ``Unknown," a designation that represents all labels unseen during training. This is  expressed as:

\begin{equation}
y(x_{t})=\left\{\begin{array}{ll}
Class & w_{t}(x_{t})>w_{0} \\
Unknown & \text { otherwise.}
\end{array}\right.
\end{equation}

\subsubsection{The Transferable Weight}
The transferable weight is derived from uncertainty and domain similarity. Similar to~\cite{you2019universal,lifshitz2020sample}, the domain similarity $d(x)$ is  obtained by the non-adversarial domain discriminator $D^{\prime}$. The $d(x)$ term  can be seen as the quantification
of the similarity of  target domain samples to the synthetic source domain samples. In particular, a smaller $d(x_{f})$ for a synthetic source sample and a larger $d(x_{t})$ for a target sample mean that they are more likely to be in the shared label set. 

On the other hand, we adopt the assumption that the target data in $Y$ have a lower uncertainty than target data in $\overline{Y_{t}}$. Thus, in order to  further separate target samples from the shared label set and those from the private label, a well-defined criterion can be used to distinguish different degrees of uncertainty. However, uncertainty is usually measured by entropy~\cite{you2019universal, saito2020universal}, which lacks discriminability for uncertainty when the categorical distributions are relatively uniform~\cite{fu2020learning}. The confidence of predicted probabilities $\bar{y}(x)$ is a better measure when  the generated categories of source samples are relatively uniform. Digging the confidence further, as private label sets of synthetic source $\overline{Y}_{f}$ have no intersection with shared label sets $Y$, samples from $p(x_{f}, y_{f} \mid y_{f} \in \overline{Y}_{f})$ are not influenced by the target data and keeps the highest certainty. In addition, the target
samples that are more similar to the source domain samples are more likely to be in the shared label set.
Different schemes of the transferable weight are further compared and analyzed in Section~\ref{Sec:model}.

With the above analysis, it is reasonable to expect that:
\begin{align}
&\mathbb{E}_{(x_{f}, y_{f}) \in p \mid y_{f} \in \overline{Y}_{f}} d(x_{f})
>\mathbb{E}_{(x_{f}, y_{f}) \in p \mid y_{f} \in Y} d(x_{f}) \nonumber\\
&~~ >\mathbb{E}_{(x_{t}, y_{t}) \in q \mid y_{t} \in Y} d(x_{t}) 
>\mathbb{E}_{(x_{t}, y_{t}) \in q \mid y_{t} \in \overline{Y}_{t}} d(x_{t}),\\
&\mathbb{E}_{(x_{f}, y_{f}) \in p \mid y_{f} \in \overline{Y}_{f}} \max \bar{y}(x_{f})
>\mathbb{E}_{(x_{f}, y_{f}) \in p \mid y_{f} \in Y} \max \overline{y}(x_{f}) \nonumber\\
&~~>\mathbb{E}_{(x_{t}, y_{t}) \in q \mid y_{t} \in Y} \max \bar{y}(x_{t}) 
>\mathbb{E}_{(x_{t}, y_{t}) \in q \mid y_{t} \in \overline{Y}_{t}} \max \bar{y}(x_{t}).
\end{align}
Thus, the sample-level  transferable weight for synthetic  source data points and target data points can be respectively defined as:
\begin{align}
w_{f}(x)&=-d(x)-\max \bar{y}(x),\\
w_{t}(x)&=d(x)+\max \bar{y}(x).
\label{eq9}
\end{align}
Note that $d(x) \in[0,1]$ and $\max \bar{y}(x) \in[0,1]$ by the max-min normalization. The weights are also
normalized into interval $[0,1]$ during training. 
\subsubsection{Domain Adaptation}
To perform domain adaptation during the training stage, the objective function aims to move the
target samples with higher transferable weight towards positive source categories $Y$. To achieve this, input $x$ from either domain
is fed into the feature extractor $F$, as shown in Fig.~\ref{fig:2}. The  extracted features $F(x)$  is forwarded into the label classifier $C$ and  the non-adversarial domain discriminator $D^{\prime}$, to obtain the transferable weights $w_{f}$ and $w_{t}$.  The  extracted feature $F(x)$  is forwarded into the  adversarial domain discriminator $D$ to adversarially align the feature distributions of
the generated source and target data falling in the shared label set. Thus,  the adversarial loss function for adaptation is defined as:
\begin{equation}
\begin{aligned}
\ell_{\mathrm{adv}}=&-\mathbb{E}_{\mathbf{x} \sim p} w_{f}(x) \log D(F(x)) \\
&-\mathbb{E}_{\mathbf{x} \sim q} w_{t}(x) \log (1-D(F(x))).
\end{aligned}
\end{equation}
Adversarially, the feature extractor $F$ strives to confuse $D$. Thus, domain-invariant features in the shared label set are obtained. In order to train the classifier $C$ on the synthetic source domain with labels, the cross-entropy loss is the following:
\begin{equation}
\ell_{\mathrm{ce}}^{f}=\mathbb{E}_{(x_{f}, y_{f}) \sim p} L( y_{f}, C(F(x_{f}))),
\end{equation}
where $L$ is the standard cross-entropy loss. Furthermore, to better reflect domain similarity, we predict samples from the synthetic source domain as 1 and samples from the target domain as 0. Thus, similar to~\cite{you2019universal,lifshitz2020sample}, a binary cross-entropy loss is used to train non-adversarial domain discriminator $D^{\prime}$.
\begin{equation}
\begin{aligned}
\ell_{\mathrm{simi}}=&-\mathbb{E}_{(x_{f}, y_{f}) \sim p} L(1, D^{\prime}(F(x_{f}))) \\
&-\mathbb{E}_{(x_{t}, y_{t}) \sim q} L(0, D^{\prime}(F(x_{t}))).
\end{aligned}
\end{equation}

\begin{center}
	\begin{algorithm}[htbp!]
		\caption{Optimization of UniDA without source data}
		\begin{algorithmic}[1]
			\REQUIRE Pre-trained model $M$ on the source domain, unlabeled data $X_{t}$ in the target domain, batch size $B$; 
			\ENSURE Classification  model $M$ of the shared classes $Y$ and the unknown class in target domain; \\
			\textbf{I. Source data generation stage:}
			\FOR{$epoch_{SDG}=1$ to $epoch_{SDG,max}$}
			\STATE Fix the pre-trained model $M$ and the style loss network (VGG-16); Randomly sample $x_{t}$ of size $B$ from $X_{t}$; 
			\FOR{each mini-batch}
			\STATE \textbf{Generate} source data $x_{f}$ by $G$, which combines  categorical vectors $y$ ($y \sim p_{y}(y)$) and standard normal vectors $z$ ($z \sim p_{z}(z)$);
			\STATE \textbf{Train} $G$ by  $\min_{\theta_{g}}(\ell_{\mathrm{cls}}(y, M(x_{f}))+\ell_{\mathrm{style }}(x_{f}, x_{t}))$;
			\ENDFOR
			\ENDFOR \\
			\textbf{II. Model adaptation stage:}\\
			\IF{starting adaptation}
			\FOR{$epoch_{MA}=1$ to $epoch_{MA,max}$}
			\STATE Randomly sample $x_{t}$ of size $B$ from $X_{t}$ and generate $x_{f}$ of size $B$ by $G$;
			\FOR{each mini-batch}
			\STATE $w_{f}$ and $w_{t}$ are obtained by $C(F(x))$ and $D^{\prime}(F(x))$;
			\STATE \textbf{Train} $D$ by  $\max_{\theta_{d}}(-\ell_{\mathrm{adv}})$;
			\STATE \textbf{Train} $F$ and $C$ by  $\min_{\theta_{f},\theta_{c}}(\ell_{\mathrm{ce}}^{f}-\ell_{\mathrm{adv}})$;
			\STATE \textbf{Train} $D^{\prime}$ by  $\min_{\theta_{d^{\prime}}}(\ell_{\mathrm{simi}})$;
			\ENDFOR
			\ENDFOR
			\ENDIF
		\end{algorithmic}
		\label{alg:1}
	\end{algorithm}
	\vspace{-5mm}
\end{center}

\subsection{Optimization}
Algorithm ~\ref{alg:1} depicts the optimization flow of UniDA without source data procedure, which consists of two independent stages. $\theta_{g}$, $\theta_{f}$, $\theta_{c}$, $\theta_{d}$, and  $\theta_{d^{\prime}}$ are parameters of $G$, $F$, $C$, $D$, and $D^{\prime}$, respectively.
First, the SDG stage estimates the conditional distribution $p(x \mid y, z)$ of source data from the pre-trained model $M$.  Thus, we train generator $G$ via the data diversity module, and combine them as a single objective function:
\begin{equation}
\ell(\theta_{g})= \min_{\theta_{g}} (\ell_{\mathrm{cls}}(y, M(x_{f}))+\ell_{\mathrm{style }}(x_{f}, x_{t})).
\end{equation}
Second, the training of the MA stage can be written as a minimax game:
\begin{align}
&\ell(\theta_{d},\theta_{f},\theta_{c}) = \max_{\theta_{d}}\min_{\theta_{f},\theta_{c}}(\ell_{\mathrm{ce}}^{f}-\ell_{\mathrm{adv}}),\\
&\ell(\theta_{d^{\prime}}) = \textstyle \min_{\theta_{d^{\prime}}}(\ell_{\mathrm{simi}}).
\end{align}
The gradient reversal layer~\cite{ganin2016domain} is used  to reverse the gradient between $F$ and $D$ to optimize the MA stage in an end-to-end training framework.
\section{Experiments}
\subsection{Experimental setup}
\subsubsection{Datasets} 

To verify our algorithm, we select the RSSCN7, UC Merced, AID, and NWPU-RESISC45 to build the cross-domain remote sensing image scene datasets. Specifically, the
\textbf{RSSCN7 dataset}~\cite{zou2015deep} contains 2800 remote sensing scene images, which are from seven typical scene categories. There are 400 images in each scene type, and each image has a size of 400$\times$400 pixels. The
\textbf{UC Merced dataset}~\cite{yang2010bag} is widely used for remote sensing image scene classification. It consists of 2100 remote sensing images from 21 scene classes. Each scene class contains 100 RGB images with an image size of 256$\times$256 pixels. The
\textbf{AID dataset}~\cite{xia2017aid} is a large-scale aerial image dataset acquired from Google Earth. It contains 10,000 images with a size of 600$\times$600 pixels, which are divided into 30 classes. The
\textbf{NWPU-RESISC45 dataset}~\cite{cheng2017remote}  consists of 31,500 remote sensing images divided into 45 scene classes. Each class includes 700 images with a size of 256×256 pixels. The spatial resolution varies from about 30 m to 0.2 m for most of the scene classes.

\begin{table*}[!h]
	\caption{Four UniDA tasks for remote sensing scene classification}
	\centering
	\resizebox{0.85\textwidth}{!}
	{\begin{tabular}{cclclclcl}
		\hline
		\hline
		RSSCN7 $\rightarrow$ UCM                  & \multicolumn{2}{c}{Total label sets}      & \multicolumn{2}{c}{Shared label sets} & \multicolumn{2}{c}{Private label   sets} & \multicolumn{2}{c}{\textbf{$\xi$}}            \\ \hline
		Source domain: RSSCN7 dataset        & \multicolumn{2}{c}{7} & \multicolumn{2}{c}{5}                 & \multicolumn{2}{c}{2}                    & \multicolumn{2}{c}{\multirow{2}{*}{0.18}} \\ \cline{1-7}
		Target domain: UC Merced dataset     & \multicolumn{2}{c}{21} & \multicolumn{2}{c}{5}                 & \multicolumn{2}{c}{16}                   & \multicolumn{2}{c}{}                      \\ \hline
		RSSCN7 $\rightarrow$ AID          & \multicolumn{2}{c}{Total label sets}              & \multicolumn{2}{c}{Shared label sets} & \multicolumn{2}{c}{Private label sets}   & \multicolumn{2}{c}{\textbf{$\xi$}}            \\ \hline
		Source domain: RSSCN7 dataset       & \multicolumn{2}{c}{7}  & \multicolumn{2}{c}{6}                 & \multicolumn{2}{c}{1}                    & \multicolumn{2}{c}{\multirow{2}{*}{0.16}} \\ \cline{1-7}
		Target domain: AID dataset           & \multicolumn{2}{c}{30}  & \multicolumn{2}{c}{6}                 & \multicolumn{2}{c}{24}                   & \multicolumn{2}{c}{}                      \\ \hline
		RSSCN7 $\rightarrow$ NWPU             & \multicolumn{2}{c}{Total label sets}           & \multicolumn{2}{c}{Shared label sets} & \multicolumn{2}{c}{Private label sets}   & \multicolumn{2}{c}{\textbf{$\xi$}}            \\ \hline
		Source domain: RSSCN7 dataset     & \multicolumn{2}{c}{7}    & \multicolumn{2}{c}{6}                 & \multicolumn{2}{c}{1}                    & \multicolumn{2}{c}{\multirow{2}{*}{0.12}} \\ \cline{1-7}
		Target domain: NWPU-RESISC45 dataset & \multicolumn{2}{c}{45} & \multicolumn{2}{c}{6}                 & \multicolumn{2}{c}{39}                   & \multicolumn{2}{c}{}                      \\ \hline
		AID $\rightarrow$ NWPU                   & \multicolumn{2}{c}{Total label sets}        & \multicolumn{2}{c}{Shared label sets} & \multicolumn{2}{c}{Private label sets}   & \multicolumn{2}{c}{\textbf{$\xi$}}            \\ \hline
		Source domain: AID dataset     & \multicolumn{2}{c}{30}      & \multicolumn{2}{c}{20}                & \multicolumn{2}{c}{10}                   & \multicolumn{2}{c}{\multirow{2}{*}{0.27}} \\ \cline{1-7}
		Target domain: NWPU-RESISC45 dataset  & \multicolumn{2}{c}{45} & \multicolumn{2}{c}{20}                & \multicolumn{2}{c}{25}                   & \multicolumn{2}{c}{}                      \\ \hline
		\hline
	\end{tabular}}
\label{Tab:Table1}
\end{table*}

\begin{figure}[t]
	\centering
	{\includegraphics[width = .5\textwidth]{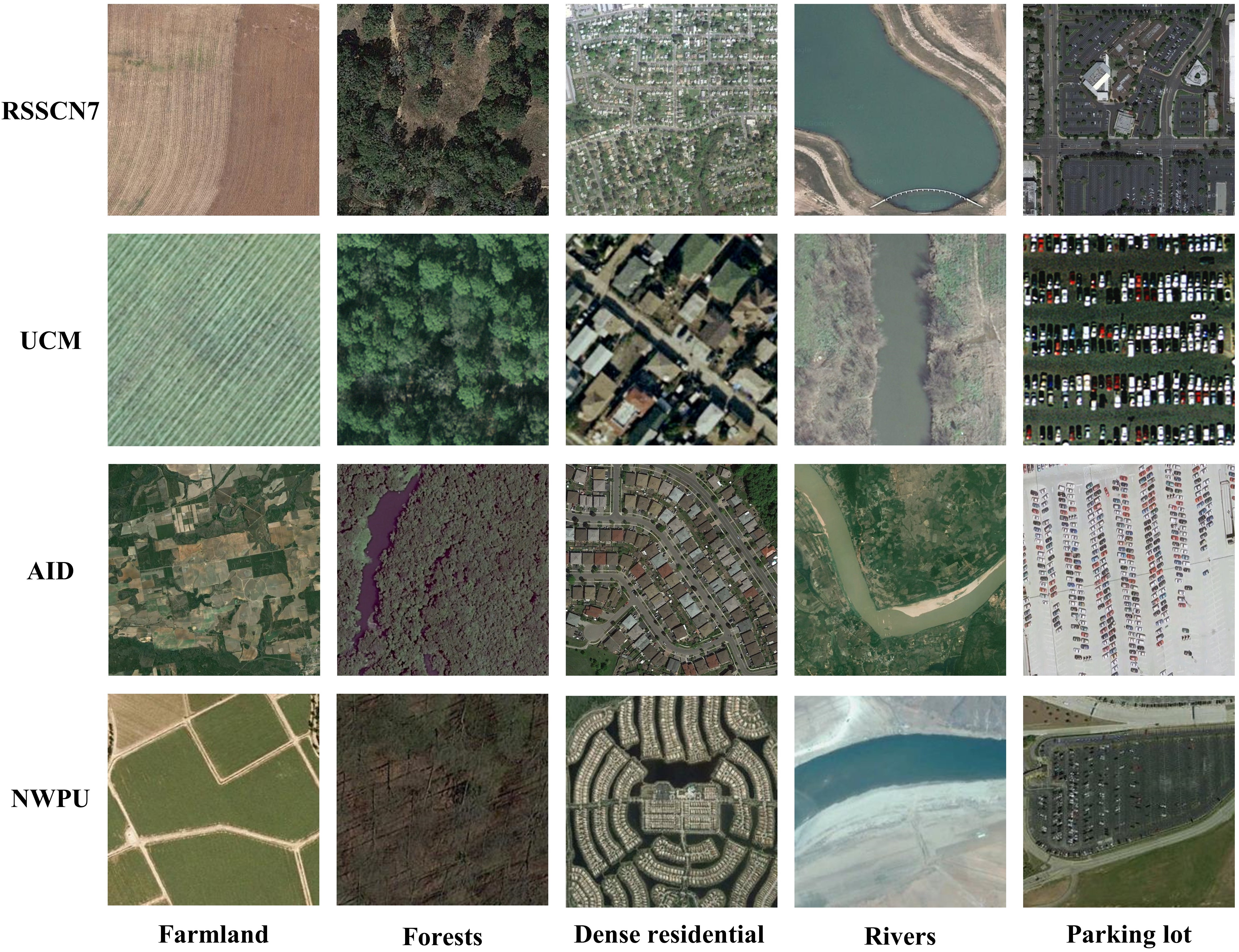}}
	\caption{ Some sample images of five shared categories extracted from four datasets. Top row to the bottom row are RSSCN7 dataset, UC Merced dataset, AID dataset, and NWPU-RESISC45 dataset, respectively.}
	\label{fig:3}
\end{figure}

As shown in Table~\ref{Tab:Table1}, four UniDA tasks for remote sensing scene classification are established. Specifically, the RSSCN7 dataset is suitable as the source domain because of its small number of categories. Thus, three cross-domain scenarios are conducted: RSSCN7 $\rightarrow$ UCM, RSSCN7 $\rightarrow$ AID, and RSSCN7 $\rightarrow$ NWPU. For  RSSCN7 $\rightarrow$ UCM, we use the five public categories as the shared label set—namely farmland, forests, dense residential areas, rivers, and parking lot—the remaining two as the private source label set, and the remaining sixteen of UC Merced as the private target label set. For RSSCN7 $\rightarrow$ AID and RSSCN7 $\rightarrow$ NWPU, we use the six public categories as the shared label set (the five previously enumerated plus industries).
In addition, a fourth, more complex UniDA task with a higher Jaccard index, AID $\rightarrow$ NWPU, is carried out. In this setting, we use the twenty public categories as the shared label set, and the rest of the AID and NWPU datasets as the private target label sets. Some sample images of shared label sets from these four datasets are shown in Fig.~\ref{fig:3}.
\subsubsection{Evaluation Protocol} 
The model is tested only on samples from the target domain; all the
target-private classes are grouped into a single ``Unknown" class. Specifically, during the testing stage of MA, if the target sample’s transferable weight is lower than a predetermined threshold $w_{0}$,  the input image is classified as ``Unknown." Thus, the average of per-class accuracy for all classes, including the shared classes and the ``Unknown" class, is the final result. Note that we run each experiment three times and report the average results.
\subsubsection{Implementation Details}
All experiments are implemented in Pytorch~\cite{paszke2017automatic}. 
In the setting of the SDG stage, we use the standard normal vector $z$ of length 10 in all experiments.  The generator $G$ is similar
to that of ACGAN~\cite{odena2017conditional}, which consists of two fully connected layers followed by seven transposed
convolutional layers (the number of convolution kernels is all four) with batch normalization after each layer.  The size of the generated image $x_{f}$ is $3 \times 256\times256$.  Adam~\cite{kingma2014adam} with a learning rate of
0.001 is used for the generator.
In addition, we compute style reconstruction loss at layers relu1$\_$2, relu2$\_$2, relu3$\_$3, and relu4$\_$3 of the VGG-16 style loss network. For the model  pre-trained on source data, it consists of a feature extractor $F$ and a classifier network $C$. A ResNet-50 model with initial weights trained on ImageNet~\cite{russakovsky2015imagenet} is used as the backbone of the feature extractor. The classifier network is a fully connected network with a single layer. The cross-entropy loss is utilized to pre-train the model on source data. The stochastic gradient descent (SGD) with a learning rate of 0.001 and momentum of 0.9 is used for the model  pre-trained on source data.
Furthermore,  the classification accuracy between the predictions of the generated data $x_{f}$ and the given label $y$ is used to compute the recoverability of the categories in the pre-trained model. After the SDG stage, the generator $G$ with the highest classification accuracy is utilized for the MA stage.

In the setting of the MA stage,  the pre-trained model from source data is used to initialize the feature extractor $F$ and the  classifier network $C$.   In addition,
The discriminators $D$ and $D^{\prime}$ consist of three fully connected layers with ReLU between the first two. We train $F$, $C$, $D$, and $D^{\prime}$  for 40000 iterations with Nesterov momentum SGD. The initial learning rate is set to 0.001, which is decayed using the same schedule as~\cite{ganin2016domain}.
During the testing stage, when the Jaccard index  $\xi \geq 0.2$ (AID $\rightarrow$ NWPU), the decision threshold $w_{0} = 0.6$. Otherwise, $w_{0}$ is set to 0.8. 
\subsubsection{Methods to Be Compared}
 We compare the performance of the proposed UniDA (with and without source data) with the following methods.
\begin{itemize}[leftmargin= 10 pt, itemsep= 0 pt, topsep = 2 pt, parsep= 2 pt]
 	\item Source-only: source-only is only trained on the real source data, and directly tested on the target domain based on the trained model and the target transferable weight.
 	\item UDA~\cite{you2019universal}: UDA is proposed to first introduce the universal DA setting in computer vision, which is a method with using source data. To discover the shared label sets and the private label sets to each domain, the transferable weights are defined based on domain similarity and entropy.
 	\item I-UAN~\cite{yin2021pseudo}: an improved universal adaptation network (I-UAN) is a UniDA method with source data. In I-UAN, the transferable weight of the source domain is defined based on a pseudo-margin vector (maximum predicted probability minus second highest predicted probability) to distinguish the shared label set. The sample-wise transferable weight of the target domain  is proposed based on the confidence to distinguish the shared and private label sets in target domain.
 	\item CMU~\cite{fu2020learning}: calibrated multiple uncertainties (CMU) is proposed, with a novel approach in which transferable weights are estimated by a mixture of complementary uncertainty quantities: entropy, confidence, and consistency. CMU is a UniDA method with real source data.
 	\item MA-only: MA-only  uses the initialized generator $G$ to generate source data. The generator is initialized randomly. Then, MA is performed between the synthetic source data and the target data. 
\end{itemize}
\subsection{Experimental Results}

\begin{table*}[t!]
	\caption{Classification accuracy of different methods on RSSCN7 $\rightarrow$ UC Merced (\%). }
	\centering
	\resizebox{1.0\textwidth}{!}
	{
			\begin{tabular}{ccccccccc}
				\bottomrule
				\multicolumn{9}{c}{\textbf{UniDA with source data}}                                                                                                                                                                                                                                                                 \\ \hline
				\multicolumn{1}{c|}{RSSCN7 $\rightarrow$ UCM}                                  & \multicolumn{6}{c|}{Category}                                                                                                                        & \multicolumn{1}{c|}{}                                       &                       \\ \cline{1-7}
				\multicolumn{1}{c|}{Methods}                                            & Farmland       & Forests        & Dense residential & Rivers         & Parking         & \multicolumn{1}{c|}{Unknown}                                & \multicolumn{1}{c|}{\multirow{-2}{*}{Avg\_Shared}}           & \multirow{-2}{*}{Avg} \\ \hline
				\multicolumn{1}{c|}{Source-only}                                       & \textbf{93.00} & 98.00          & 42.00             & 60.00          & 67.00           & \multicolumn{1}{c|}{1.63}                                   & \multicolumn{1}{c|}{72.00}                                  & 60.27                 \\ \hline
				\multicolumn{1}{c|}{I-UAN~\cite{yin2021pseudo}}                                             & 92.00          & 97.00          & 72.00             & 70.00          & 30.00           & \multicolumn{1}{c|}{11.63}                                  & \multicolumn{1}{c|}{72.20}                                  & 62.10                 \\ 
				\multicolumn{1}{c|}{UDA~\cite{you2019universal}}                                               & 87.00          & 98.00          & \textbf{83.00}    & \textbf{84.00} & 68.00           & \multicolumn{1}{c|}{1.13}                                   & \multicolumn{1}{c|}{84.00}                                  & 70.19                 \\ 
				\multicolumn{1}{c|}{CMU~\cite{fu2020learning}}                                               & 89.00          & \textbf{99.00} & 68.00             & 78.00          & \textbf{100.00} & \multicolumn{1}{c|}{12.25}                                  & \multicolumn{1}{c|}{86.80}                                  & 74.38                 \\ \hline
				\multicolumn{1}{c|}{MA}    & 85.00          & 98.00          & 74.00             & 79.00          & \textbf{100.00} & \multicolumn{1}{c|}{\textbf{14.63}} & \multicolumn{1}{c|}{\textbf{87.20}} & \textbf{75.11}        \\ \hline
				\multicolumn{9}{c}{\textbf{UniDA without source data}}                                                                                                                                                                                                                                                              \\ \hline
				\multicolumn{1}{c|}{MA-only}                             & 77.00          & 13.00          & 55.00             & 69.00          & 77.00           & \multicolumn{1}{c|}{0.44}                                   & \multicolumn{1}{c|}{58.20}                                  & 48.57                 \\ \hline
				\multicolumn{1}{c|}{SDG-MA w/o d}                   & 60.00          & 55.00          & 31.00             & \textbf{86.00} & 86.00           & \multicolumn{1}{c|}{4.38}                                   & \multicolumn{1}{c|}{63.60}                                  & 53.73                 \\ 
				\multicolumn{1}{c|}{SDG-MA w/o y}                   & 90.00          & 56.00          & 28.00             & 70.00          & \textbf{100.00} & \multicolumn{1}{c|}{0.19}                                   & \multicolumn{1}{c|}{68.80}                                  & 57.36                 \\ \hline
				\multicolumn{1}{c|}{SDG-MA} & \textbf{92.00} & \textbf{64.00} & \textbf{63.00}    & 76.00          & 93.00           & \multicolumn{1}{c|}{\textbf{17.13}} & \multicolumn{1}{c|}{\textbf{77.60}} & \textbf{67.52}        \\ \hline
			\end{tabular}}
\label{Tab:Table2}
\end{table*}

\begin{table*}[t!]
	\caption{Classification accuracy of different methods on RSSCN7 $\rightarrow$ AID (\%).}
	\centering
	\resizebox{1.0\textwidth}{!}
	{
	\begin{tabular}{cccccccccc}
		\bottomrule
		\multicolumn{10}{c}{\textbf{UniDA with source data}}                                                                                                                                                                                                                            \\ \hline
		\multicolumn{1}{c|}{RSSCN7 $\rightarrow$ AID}                      & \multicolumn{7}{c|}{Category}                                                                                                                 & \multicolumn{1}{c|}{\multirow{2}{*}{Avg\_Shared}} & \multirow{2}{*}{Avg} \\ \cline{1-8}
		\multicolumn{1}{c|}{Mehods}                           & Farmland       & Forests         & Dense residential & Rivers         & Parking        & Industries     & \multicolumn{1}{c|}{Unknown}        & \multicolumn{1}{c|}{}                             &                      \\ \hline
		\multicolumn{1}{c|}{Source-only}                     & 79.19          & \textbf{100.00} & 90.73             & 63.41          & 80.00          & 68.46          & \multicolumn{1}{c|}{0.12}           & \multicolumn{1}{c|}{80.30}                        & 68.84                \\ \hline
		\multicolumn{1}{c|}{I-UAN~\cite{yin2021pseudo}}                           & 93.24          & 99.60           & 94.63             & 59.27          & 81.79          & 61.79          & \multicolumn{1}{c|}{10.84}          & \multicolumn{1}{c|}{81.72}                        & 71.60                \\
		\multicolumn{1}{c|}{UDA~\cite{you2019universal}}                             & 92.16          & 99.60           & \textbf{97.56}    & 47.56          & 93.33          & 68.21          & \multicolumn{1}{c|}{2.52}           & \multicolumn{1}{c|}{83.07}                        & 71.56                \\
		\multicolumn{1}{c|}{CMU~\cite{fu2020learning}}                             & 83.51          & 98.00           & 89.51             & \textbf{59.51} & 91.79          & \textbf{80.77} & \multicolumn{1}{c|}{10.51}          & \multicolumn{1}{c|}{83.85}                        & 73.37                \\ \hline
		\multicolumn{1}{c|}{MA}          & \textbf{92.70} & \textbf{100.00} & 94.63             & 58.78          & \textbf{94.36} & 70.51          & \multicolumn{1}{c|}{\textbf{13.48}} & \multicolumn{1}{c|}{\textbf{85.16}}               & \textbf{74.92}       \\ \hline
		\multicolumn{10}{c}{\textbf{UniDA without source data}}                                                                                                                                                                                                                         \\ \hline
		\multicolumn{1}{c|}{MA-only}           & 91.89          & 16.00           & 55.37             & 73.41          & 56.67          & 59.49          & \multicolumn{1}{c|}{0.36}           & \multicolumn{1}{c|}{58.81}                        & 50.46                \\ \hline
		\multicolumn{1}{c|}{SDG-MA w/o d} & 92.70          & \textbf{93.60}  & 61.22             & 47.32          & \textbf{99.49} & 56.92          & \multicolumn{1}{c|}{2.92}           & \multicolumn{1}{c|}{\textbf{75.21}}               & 64.88                \\
		\multicolumn{1}{c|}{SDG-MA w/o y} & 90.00          & 67.20           & 71.22             & \textbf{77.56} & 92.31          & 35.38          & \multicolumn{1}{c|}{6.11}           & \multicolumn{1}{c|}{72.28}                        & 62.83                \\ \hline
		\multicolumn{1}{c|}{SDG-MA}       & \textbf{97.84} & 69.20           & \textbf{78.78}    & 37.32          & 96.92          & \textbf{62.05} & \multicolumn{1}{c|}{\textbf{17.93}} & \multicolumn{1}{c|}{73.69}                        & \textbf{65.72}       \\ \hline
	\end{tabular}}
\label{Tab:Table3}
\end{table*}

\begin{table*}[t!]
	\caption{Classification accuracy of different methods on RSSCN7 $\rightarrow$ NWPU-RESISC45 (\%).}
	\centering
	\resizebox{1.0\textwidth}{!}
	{\begin{tabular}{cccccccccc}
		\bottomrule
		\multicolumn{10}{c}{\textbf{UniDA with source data}}                                                                                                                                                                                                                     \\ \hline
		\multicolumn{1}{c|}{RSSCN7 $\rightarrow$ NWPU}               & \multicolumn{7}{c|}{Category}                                                                                                                & \multicolumn{1}{c|}{\multirow{2}{*}{Avg\_Shared}} & \multirow{2}{*}{Avg} \\ \cline{1-8}
		\multicolumn{1}{c|}{Methods}                   & Farmland       & Forests        & Dense residential & Rivers         & Parking        & Industries     & \multicolumn{1}{c|}{Unknown}        & \multicolumn{1}{c|}{}                             &                      \\ \hline
		\multicolumn{1}{c|}{Source-only}               & 70.14          & 97.71          & 67.71             & 34.14          & 74.00          & 74.00          & \multicolumn{1}{c|}{0.23}           & \multicolumn{1}{c|}{69.62}                        & 59.71                \\ \hline
		\multicolumn{1}{c|}{I-UAN~\cite{yin2021pseudo}}                     & \textbf{85.71} & \textbf{98.14} & \textbf{96.43}    & 35.57          & 28.43          & \textbf{89.00} & \multicolumn{1}{c|}{5.52}           & \multicolumn{1}{c|}{72.21}                        & 62.69                \\
		\multicolumn{1}{c|}{UDA~\cite{you2019universal}}                       & 77.29          & 97.29          & 88.14             & 40.86          & 90.86          & 76.71          & \multicolumn{1}{c|}{2.93}           & \multicolumn{1}{c|}{78.52}                        & 67.72                \\
		\multicolumn{1}{c|}{CMU~\cite{fu2020learning}}                       & 80.00          & 95.57          & 72.86             & \textbf{47.71} & 84.14          & 80.43          & \multicolumn{1}{c|}{12.18}          & \multicolumn{1}{c|}{76.79}                        & 67.56                \\ \hline
		\multicolumn{1}{c|}{MA}    & 84.57          & 94.29          & 88.43             & 40.00          & \textbf{92.00} & 80.14          & \multicolumn{1}{c|}{\textbf{14.36}} & \multicolumn{1}{c|}{\textbf{79.90}}               & \textbf{70.54}       \\ \hline
		\multicolumn{10}{c}{\textbf{UniDA without source data}}                                                                                                                                                                                                                  \\ \hline
		\multicolumn{1}{c|}{MA-only}     & 83.29          & 0.00           & 10.00             & 0.00           & 36.57          & 24.00          & \multicolumn{1}{c|}{22.04}          & \multicolumn{1}{c|}{25.64}                        & 25.13                \\ \hline
		\multicolumn{1}{c|}{SDG-MA w/o d} &91.43 	&37.29 	&78.57 	&\textbf{59.71} 	&76.14 	&\textbf{76.57} 	& \multicolumn{1}{c|}{1.21} 	 & \multicolumn{1}{c|}{69.95} 	& 60.13 \\
		\multicolumn{1}{c|}{SDG-MA w/o y} &\textbf{93.14} 	&52.14 	&39.43 	&37.86 	&81.86 	&42.43 	&\multicolumn{1}{c|}{12.94} 	& \multicolumn{1}{c|}{57.81} 	&51.40 
		 \\  \hline
		\multicolumn{1}{c|}{SDG-MA} & 91.86 & \textbf{88.00} & \textbf{86.57}    & 28.00 & \textbf{84.86} & 65.86 & \multicolumn{1}{c|}{\textbf{25.20}} & \multicolumn{1}{c|}{\textbf{74.19}}               & \textbf{67.19}       \\ \hline
	\end{tabular}}
\label{Tab:Table4}
\end{table*}

\begin{table*}[t!]
	\caption{Classification accuracy of different methods on AID$\rightarrow$NWPU (\%). Fa.: Farmland, Fo.: Forest, DR: Dense residential, Ri.: River, Pa.: Parking, In.: Industrial, Be.: Beach, MR: Medium residential, SR: Sparse residential, Ai.: Airport, Br.: Bridge, Ba.: BaseballField, Ch.: Church, De.: Desert, Me.: Meadow, Mo.: Mountain, RS: Railway station, St.: Stadium, ST: Storage tanks, Co.: Commercial.}
	\centering
	\resizebox{1.06\textwidth}{!}
	{\begin{tabular}{cccccccccccccccccccccccc}
				\bottomrule
				\multicolumn{24}{c}{\textbf{UniDA with source data}}                                                                           \\ \hline
				\multicolumn{1}{c|}{AID$\rightarrow$NWPU}                          & \multicolumn{21}{c|}{Category}   & \multicolumn{1}{l|}{\multirow{2}{*}{Avg\_Shared}} & \multicolumn{1}{l}{\multirow{2}{*}{Avg}} \\ \cline{1-22}
				\multicolumn{1}{c|}{Methods}                           & Fa.       & Fo.         & DR & Ri.          & Pa.        & In.     & Be.          & MR & SR & Ai.        & Br.         & Ba.  & Ch.         & De.         & Me.         & Mo.       & RS & St.        & ST  & Co.     & \multicolumn{1}{c|}{Unknown}        & \multicolumn{1}{l|}{}                             & \multicolumn{1}{l}{}                     \\ \hline
				\multicolumn{1}{c|}{Source-only}                       & 89.00          & 86.00          & 50.71             & 66.00          & 78.14          & 16.57          & 0.00           & 66.71              & 78.86              & 62.43          & 0.00           & 0.00           & 78.43          & 89.86          & 75.43          & 93.43          & 0.00            & 0.00           & 93.57          & 54.57          & \multicolumn{1}{c|}{0.01}           & \multicolumn{1}{c|}{53.99}                        & 51.42                                    \\ \hline
				\multicolumn{1}{c|}{I-UAN~\cite{yin2021pseudo}}                             & \textbf{96.00} & \textbf{96.29} & \textbf{78.86}    & 74.86          & 86.29          & 75.86          & 94.86          & \textbf{83.00}     & 86.57              & 51.00          & 90.29          & 78.71          & \textbf{83.43} & \textbf{80.71} & 88.43          & 81.71          & 84.14           & \textbf{81.00} & \textbf{95.29} & 57.00          & \multicolumn{1}{c|}{10.52}          & \multicolumn{1}{c|}{\textbf{82.21}}               & \textbf{78.80}                           \\
				\multicolumn{1}{c|}{UDA~\cite{you2019universal}}                               & 91.71          & 92.57          & 78.29             & 71.29          & 81.14          & 77.29          & 91.57          & 70.00              & 83.00              & 45.14          & 81.29          & 74.57          & 73.29          & 78.71          & 80.14          & 85.43          & 82.14           & 50.00          & 88.43          & 37.43          & \multicolumn{1}{c|}{2.57}           & \multicolumn{1}{c|}{75.67}                        & 72.19                                    \\
				\multicolumn{1}{c|}{CMU~\cite{fu2020learning}}                               & 89.57          & 83.86          & 80.43             & 76.43          & 84.00          & 71.00          & 88.57          & 79.29              & 87.00              & 39.00          & 86.71          & 73.29          & 74.00          & 72.00          & 75.57          & \textbf{89.71} & \textbf{87.43}  & 60.43          & 94.00          & 54.14          & \multicolumn{1}{c|}{13.81}          & \multicolumn{1}{c|}{77.32}                        & 74.30                                    \\ \hline
				\multicolumn{1}{c|}{MA}        & 93.86          & 87.43          & 64.14             & \textbf{78.57} & \textbf{89.00} & \textbf{83.43} & \textbf{95.43} & 82.57              & \textbf{88.00}     & \textbf{58.00} & \textbf{94.71} & \textbf{79.00} & 76.29          & 79.86          & \textbf{91.57} & 70.71          & 84.86           & 46.00          & 92.86          & \textbf{60.14} & \multicolumn{1}{c|}{\textbf{14.11}} & \multicolumn{1}{c|}{79.82}                        & 76.69                                    \\ \hline
				\multicolumn{24}{c}{\textbf{UniDA without source data}}                                                                                           \\ \hline
				\multicolumn{1}{c|}{MA-only}             & 0.29           & 0.00           & 0.00              & 0.14           & 6.29           & 0.00           & 6.71           & 0.00               & 0.00               & 14.86          & 4.14           & 0.00           & 0.00           & 0.00           & 0.00           & 0.00           & 1.86            & 0.00           & 85.14          & 0.00           & \multicolumn{1}{c|}{30.83}          & \multicolumn{1}{c|}{5.97}                         & 7.16                                     \\ \hline
				\multicolumn{1}{c|}{SDG-MA w/o   d} & 70.71          & 63.29          & \textbf{51.71}    & 62.57          & \textbf{85.57} & 81.29          & 79.71          & 22.57              & 81.29              & 22.14          & 86.71          & \textbf{83.29} & 36.29          & 88.86          & \textbf{89.43} & 3.43           & 78.43           & \textbf{80.86} & 80.86          & 39.43          & \multicolumn{1}{c|}{4.03}           & \multicolumn{1}{c|}{64.42}                        & 61.55                                    \\
				\multicolumn{1}{c|}{SDG-MA w/o   y} & \textbf{82.57} & 2.57           & 29.86             & 70.43          & 84.57          & 24.71          & \textbf{95.71} & \textbf{44.43}     & 76.14              & \textbf{64.57} & 92.71          & 75.57          & \textbf{62.71} & 83.29          & 76.43          & \textbf{20.71} & \textbf{82.14}  & 69.86          & \textbf{90.29} & 31.29          & \multicolumn{1}{c|}{8.07}           & \multicolumn{1}{c|}{63.03}                        & 60.41                                    \\ \hline
				\multicolumn{1}{c|}{SDG-MA}         & 80.57          & \textbf{79.43} & 29.57             & \textbf{77.00} & 83.71          & \textbf{82.29} & 79.57          & 20.57              & \textbf{90.43}     & 44.86          & \textbf{94.00} & 81.00          & 52.86          & \textbf{91.71} & 88.57          & 12.00          & 73.00           & 62.71          & 78.71          & \textbf{41.57} & \multicolumn{1}{c|}{\textbf{20.22}} & \multicolumn{1}{c|}{\textbf{67.21}}               & \textbf{64.97}                           \\ \hline
			\end{tabular} }
		\label{Tab:Table5}
\end{table*}

\begin{figure*}[htp!]
	\centering
	{\includegraphics[width = 0.8\textwidth]{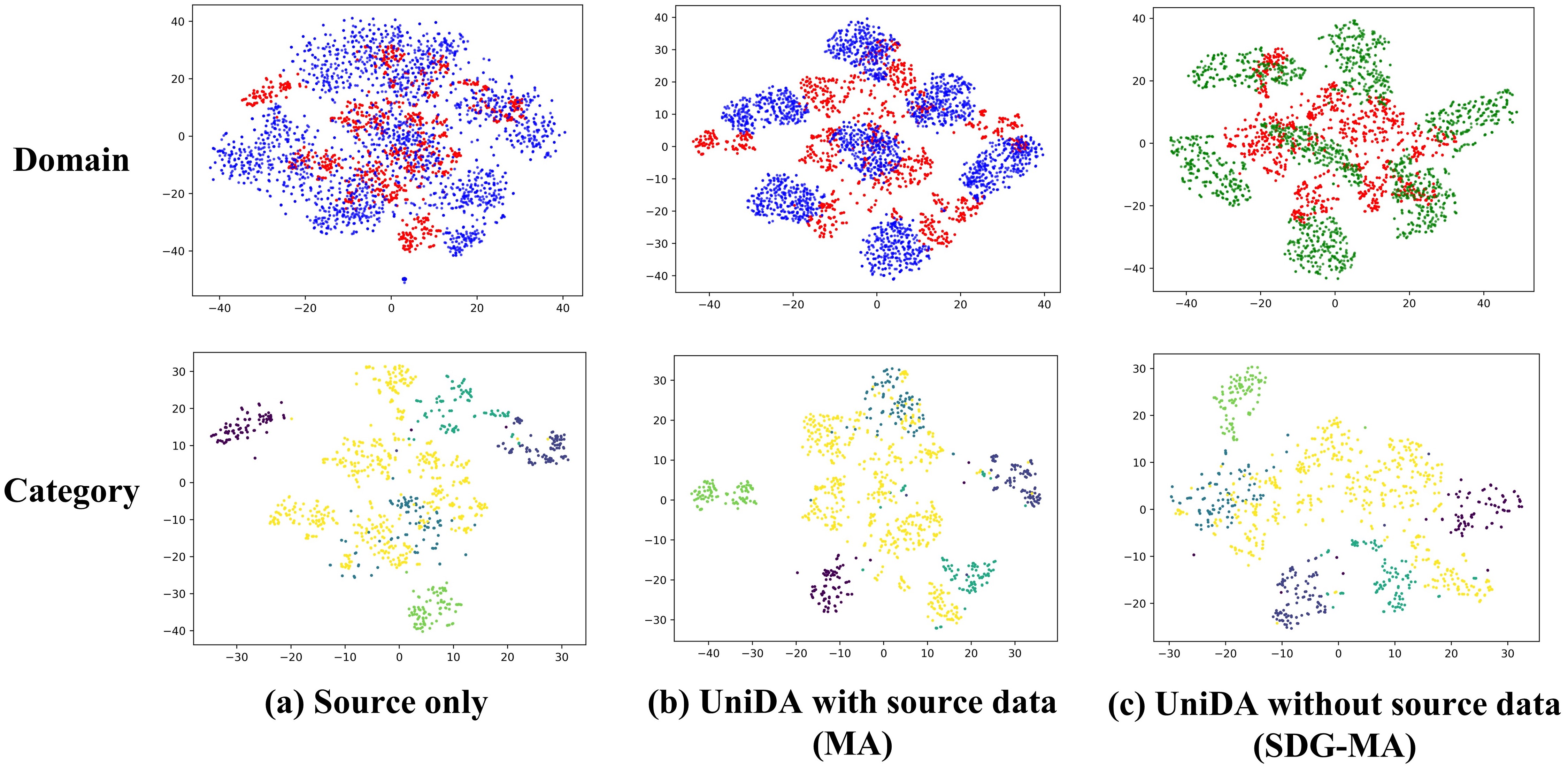}}
	\vspace{-4mm}
	\caption{ Feature visualization on RSSCN7 $\rightarrow$ UC Merced. For domain, blue and green represents the real source domain and synthetic source domain, respectively. Red refers to the target domain. For category, yellow plots are ``unknown'' samples, others are ``known" samples.}
	\label{fig:4}
\end{figure*}

\begin{figure*}[htp!]
	\centering
	{\includegraphics[width = 0.8\textwidth]{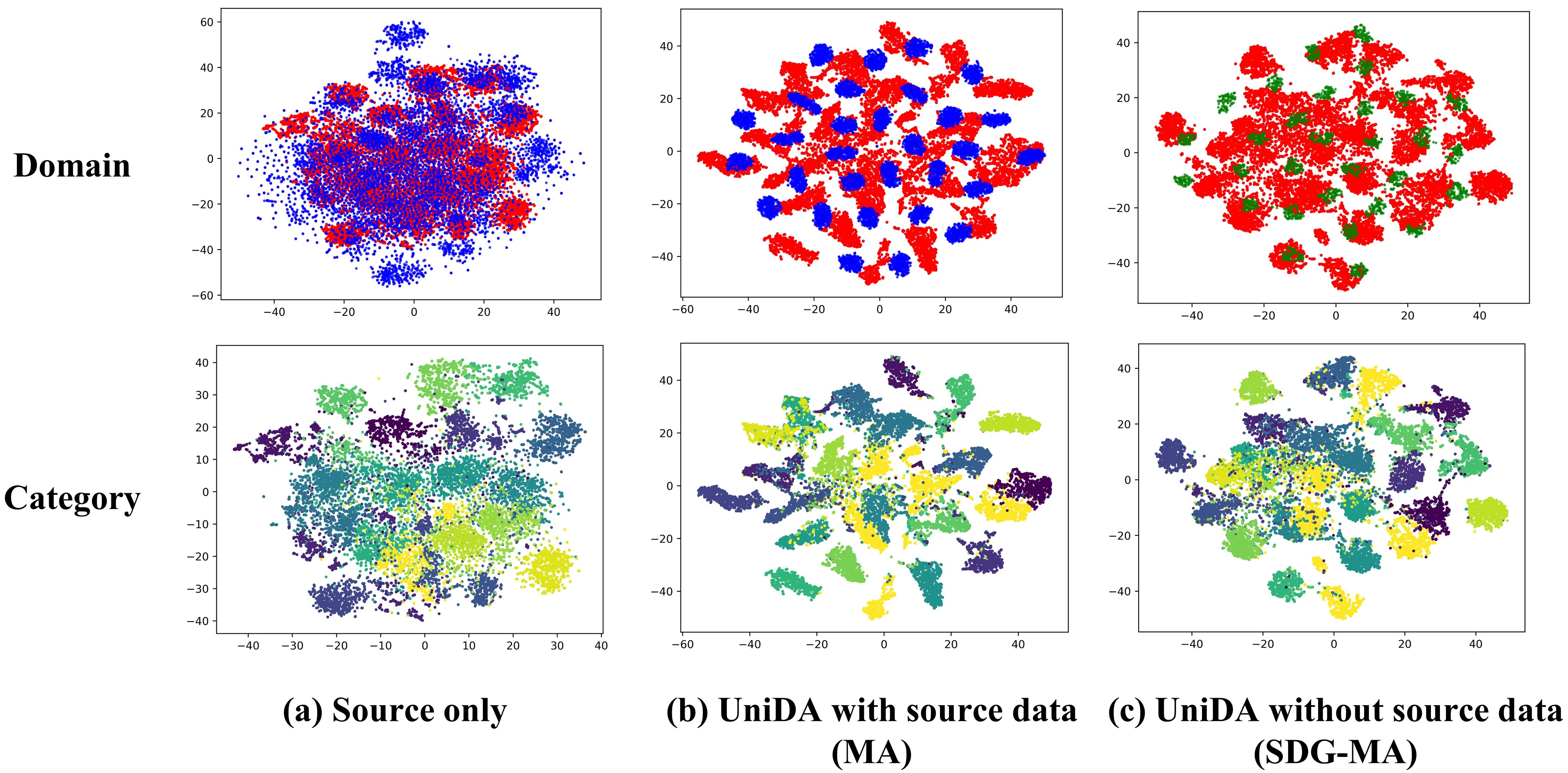}}
	\vspace{-4mm}
	\caption{Feature visualization on AID $\rightarrow$ NWPU. For domain, blue and green represents the real source domain and synthetic source domain, respectively. Red refers to the target domain. For category, yellow plots are ``unknown" samples, others are ``known" samples.}
	\label{fig:5}
\end{figure*}

\begin{figure*}[htp!]
	\centering
	\subfigure[\textcolor{black}{Threshold parameter $w_{0}$ on RSSCN7 $\rightarrow$ UC Merced}]{
		\includegraphics[width = .3\linewidth]{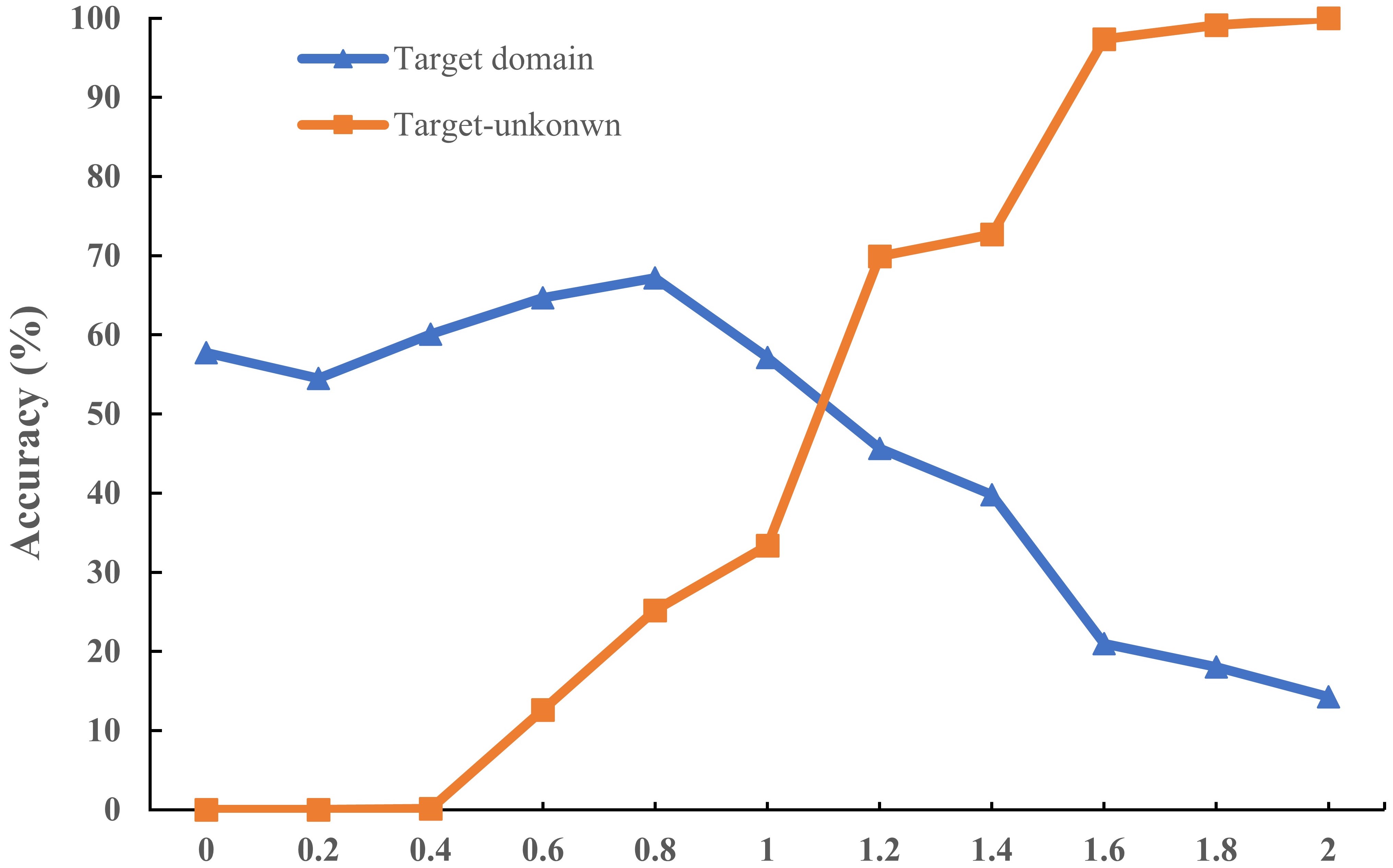}
	}
	\subfigure[\textcolor{black}{Threshold parameter $w_{0}$ on AID $\rightarrow$ NWPU}]{
		\includegraphics[width = .3\linewidth]{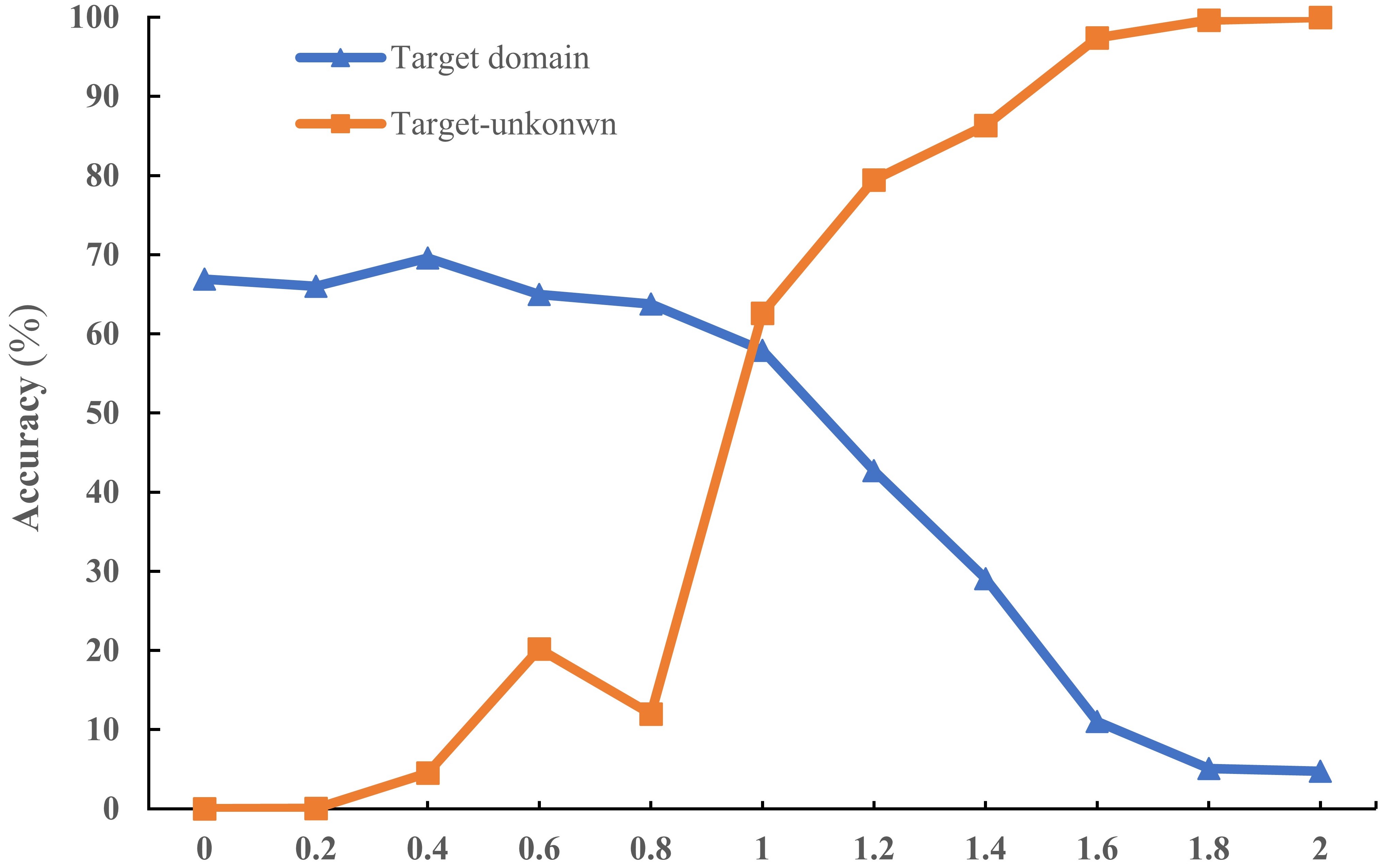}
	}
        \subfigure[\textcolor{black}{Size of shared label sets on AID $\rightarrow$ NWPU}]{
		\includegraphics[width = .26\linewidth]{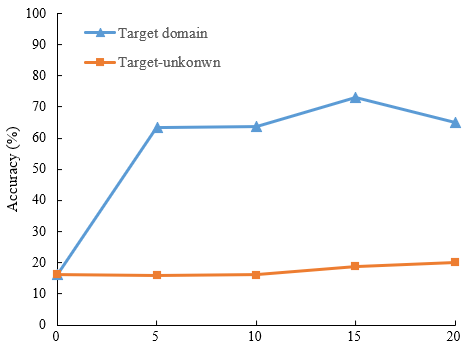}
	}
	\caption{\textcolor{black}{Decision threshold analysis for the parameters $w_{0}$ and varying size of shared label sets $Y$.}}
	\label{fig:6}
\end{figure*}

\subsubsection{Results on RSSCN7 $\rightarrow$ UCM} Our first experiment is conducted on RSSCN7 $\rightarrow$ UCM, including two cases using UniDA with source data and UniDA without source data. The results are listed in Table~\ref{Tab:Table2}. From Table~\ref{Tab:Table2}, in the UniDA setting with source data, it can be seen that the accuracy of all methods improves compared to source-only. This phenomenon illustrates that a domain shift appears in RSSCN7 and UCM datasets. Furthermore, in the  UniDA with source data setting, our proposed MA achieves much better performance than all the other baselines, with an average accuracy of 75.11\%. In particular, the average accuracy of shared label sets and all label sets improves by 0.4\% and 0.73\%, respectively,  compared with the best baseline CMU~\cite{fu2020learning}. These findings demonstrate that the proposed sample-level transferable weight in MA, including  confidence and domain similarity, is more efficient than entropy in  UDA~\cite{you2019universal}, pseudo-margin vector in I-UAN~\cite{yin2021pseudo}, and the mixture of entropy, confidence, and consistency in CMU~\cite{fu2020learning} for remote sensing image scene classification. 

In the second case of UniDA without source data, we observe that our proposed SDG-MA framework significantly outperforms the MA-only method by 18.95\% on the average accuracy of all label sets. It is obvious that the proposed source data generation is effective
and practical in the UniDA setting without source data of remote sensing images.
Notably, compared with the MA, our SDG-MA maintains a more prominent performance in the ``Unknown" class. It has been demonstrated that data points generated by the SDG effectively cover the distribution of the source data.

\subsubsection{Results on RSSCN7 $\rightarrow$ AID} Our second experiment is conducted on RSSCN7 $\rightarrow$ AID; results are shown in Table~\ref{Tab:Table3}. A similar tendency is observed in the UniDA with source data setting for remote sensing images. The proposed  MA outperforms all the compared methods. Again, the experiment demonstrates that the proposed sample-level transferable weight filters out data coming from shared and private label sets on feature alignment and provides a better criterion for ``unknown" class detection than the existing methods. In addition, the MA achieves the best accuracy outcomes compared with the baselines, for some shared categories (such as ``Farmland," ``Forests," and ``Parking"). 

In the UniDA setting without source data, our proposed method has improved by 15.26\% compared with the MA-only method. This phenomenon once again verifies the reliability of the proposed SDG. For identifying unknown classes, the SDG-MA  yields excellent performance. However, the average accuracy of all categories is lower than the source-only method. The reason for this finding is that the difference between RSSCN7 and AID in the shared label sets is relatively smaller than other UniDA tasks.

\subsubsection{Results on RSSCN7 $\rightarrow$ NWPU} Our third experiment is conducted on the RSSCN7 $\rightarrow$ NWPU, and the results are provided in Table~\ref{Tab:Table4}. In the UniDA setting with source data for remote sensing images, our proposed source-base UniDA is 10.83  percentage points greater than the source-only method and achieves the highest average accuracy among all methods for recognizing all target samples, which verifies the effectiveness of our proposed MA. 

In the UniDA setting without source data, our proposal improves by 42.06\%, compared with the MA-only method. Notably, the SDG-MA method exhibits a huge performance for ``Unknown" category.

\subsubsection{Results on AID $\rightarrow$ NWPU} The experimental results on AID $\rightarrow$ NWPU are reported in Table~\ref{Tab:Table5}. In the UniDA setting with source data, the proposed MA improves by 25.27\% compared with the source-only method, which confirms the effectiveness and practicality of the proposed MA in a complex UniDA task with a higher Jaccard index. In addition, our proposed method achieves superior performance among all methods on most shared categories and can achieve the highest classification accuracy among all methods for recognizing private label sets in target samples (the ``Unknown" category). However, I-UAN~\cite{yin2021pseudo} is better than MA in identifying shared categories, because the proposed MA has a significant drop in the accuracy of some shared categories, such as ``Stadium."

In the UniDA setting without source data, the MA-only method achieves poor results (only 7.16\%) due to a lack of reliable source domain data. Conversely, SDG-MA achieves superior performance (64.97\%) because reliable source data is generated by SDG. Furthermore, SDG-MA achieves the highest accuracy for identifying the ``Unknown" category, which further proves that the proposed SDG module can generate a uniform distribution that approximates the real source domain.

\subsection{Model Analysis \label{Sec:model}}
\subsubsection{Feature Distribution Analysis}
To fully understand the proposed UniDA with source data and UniDA without source data, we provide the feature distributions of RSSCN7 $\rightarrow$ UCM and AID $\rightarrow$ NWPU in Figs.~\ref{fig:4} and~\ref{fig:5}, respectively. The t-SNE~\cite{van2008visualizing} is used to visualize the learned source and target features with corresponding domain labels and category labels. As shown in Fig.~\ref{fig:4}(a),  before adaptation (source only),  there are domain shifts between the real source domain (blue) and the target domain (red) according to the domain distribution. From the category labels, the distributions of the shared categories are fragmented and most target private samples  are attached near the shared samples. After applying MA (Fig.~\ref{fig:4}(b)) and SDG-MA (Fig.~\ref{fig:4}(c)), domain shifts are effectively alleviated. In addition, separability between shared categories is increased  and most target private samples are separated from the shared samples.
These phenomena demonstrate that our proposed MA strategy is effective for feature alignment. Furthermore, comparing MA (Fig.~\ref{fig:4}(b)) and SDG-MA  (Fig.~\ref{fig:4}(c)), we can observe that the synthetic source data distribution (green) and the real source data distribution (blue) show a high degree of consistency in class distribution and data diversity. It has been demonstrated that the synthetic source data generated by SDG is effective and reliable.

In Fig.~\ref{fig:5}(a),  we can observe that the real source domain and the target domain have larger data shifts in the UniDA task AID $\rightarrow$ NWPU than the task RSSCN7 $\rightarrow$ UCM. After applying MA  (Fig.~\ref{fig:5}(b)), it is clear that MA alleviates the distribution discrepancy in domain labels. Again, this finding  demonstrates that the proposed MA is practical and effective. After applying SDG-MA  (Fig.~\ref{fig:5}(c)), intra-class compactness and inter-class separability are significantly improved compared with source only. In addition, the intra-class compactness of the ``Unknown" category (Fig.~\ref{fig:5}(c)) is improved compared with MA (Fig.~\ref{fig:5}(b)). This phenomenon further verifies the validity of the generated source data.

\begin{table*}[htp!]
	\caption{Analysis of source data  generation  on RSSCN7 $\rightarrow$ UC Merced.}
	\centering
	\resizebox{.85\textwidth}{!}
	{
		\begin{tabular}{cccccccccc}
\hline
\multirow{2}{*}{Stages}  & \multirow{2}{*}{Methods}   & \multicolumn{8}{c}{RSSCN7 $\rightarrow$   UC Merced}                                                                                     \\ \cline{3-10} 
&                            & Farmland       & Forests        & Dense residential & Rivers         & Parking        & Unknown        & Avg\_Shared    & Avg            \\ \hline
\multirow{2}{*}{SDG}   & SDG-MA w/o classifier loss & 2.00           & 2.00           & 19.00             & 4.00           & 17.00          & 10.38          & 8.80           & 9.06           \\
& SDG-MA w/o style loss      & 85.00          & 0.00           & 0.00              & 7.00           & 81.00          & 16.00          & 34.60          & 31.50          \\ \hline
\multirow{3}{*}{\begin{tabular}[c]{@{}c@{}}Different Data Diversity \\   Generation   Schemes\end{tabular}} & 3C-GAN~\cite{li2020model}                     & 91.00          & 1.00           & 1.00              & 9.00           & 87.00          & 15.31          & 37.80          & 34.05          \\
 & KEGNET~\cite{yoo2019knowledge}                     & 31.00          & 12.00          & 21.00             & \textbf{82.00} & 72.00          & \textbf{23.13} & 43.60          & 40.19          \\
 & Ours                       & \textbf{92.00} & \textbf{64.00} & \textbf{63.00}    & 76.00          & \textbf{93.00} & 17.13          & \textbf{77.60} & \textbf{67.52} \\ \hline
\end{tabular}}
	\label{Tab:Table6}
\end{table*}

\begin{table*}[t!]
	\caption{Analysis of source data  generation on AID$\rightarrow$NWPU. Fa.: Farmland, Fo.: Forest, DR: Dense residential, Ri.: River, Pa.: Parking, In.: Industrial, Be.: Beach, MR: Medium residential, SR: Sparse residential, Ai.: Airport, Br.: Bridge, Ba.: BaseballField, Ch.: Church, De.: Desert, Me.: Meadow, Mo.: Mountain, RS: Railway station, St.: Stadium, ST: Storage tanks, Co.: Commercial.}
	\centering
	\resizebox{1.05\textwidth}{!}
	{
		\begin{tabular}{ccccccccccccccccccccccccc}
			\hline
			\multirow{2}{*}{Stages} & \multirow{2}{*}{Methods} & \multicolumn{23}{c}{AID$\rightarrow$NWPU}                                                                     \\ \cline{3-25} 
		& 	& Fa.       & Fo.         & DR & Ri.          & Pa.        & In.     & Be.          & MR & SR & Ai.        & Br.         & Ba.  & Ch.         & De.         & Me.         & Mo.       & RS & St.        & ST  & Co.     & Unknown        & Avg\_Shared    & Avg            \\ \hline
  
 \multirow{2}{*}{SDG}   &  SDG-MA   w/o classifier loss & 0.00                 & 0.00                 & 11.71                & 0.00                 & 0.00                 & 7.71                 & 0.00                 & 19.43                & 0.00                 & 20.00                & 0.00                 & 1.43                 & 8.71                 & 0.00                 & 0.00                 & 0.00                 & 1.14                 & 0.00                 & 5.14                 & 1.29                 & \textbf{39.93}               & 3.83                 & 5.55                 \\
& SDG-MA w/o style loss        & 65.57                & 35.71                & 7.14                 & 36.57                & 76.14                & 63.29                & 55.57                &  \textbf{46.57}               & 71.14                & 25.57                & 93.57                & 78.86                & 47.14                & 87.43                & 82.71                & 0.14                 & \textbf{75.86}                & 6.71                 & 53.43                & 50.43                & 15.81                & 52.98                & 51.21                \\ \hline
			\multirow{3}{*}{\begin{tabular}[c]{@{}c@{}}Different Data Diversity \\    Generation  Schemes\end{tabular}} & 3C-GAN~\cite{li2020model}         & 69.43          & 71.00          & 19.43             & 57.43          & 82.86          & 71.00          & 78.00          & 45.14     & 80.71              & 30.43          & 90.57          & 82.14          & 42.86          & \textbf{93.43} & 76.71          & 0.00           & 73.43  & 61.29          & \textbf{84.71} & \textbf{50.57} & 22.77 & 63.06          & 61.14          \\
		&	KEGNET~\cite{yoo2019knowledge}             & 53.29          & 16.57          & 16.57             & 35.57          & \textbf{87.00} & 64.43          & 29.86          & 25.00              & 35.29              & 14.14          & 75.29          & \textbf{83.00} & 39.86          & 87.43          & 65.43          & 0.00           & 16.43           & 18.43          & 43.71          & 38.57          & 19.30          & 42.29          & 41.20          \\
			& Ours       & \textbf{80.57} & \textbf{79.43} & \textbf{29.57}    & \textbf{77.00} & 83.71          & \textbf{82.29} & \textbf{79.57} & 20.57              & \textbf{90.43}     & \textbf{44.86} & \textbf{94.00} & 81.00          & \textbf{52.86} & 91.71          & \textbf{88.57} & \textbf{12.00} & 73.00           & \textbf{62.71} & 78.71          & 41.57          & 20.22          & \textbf{67.21} & \textbf{64.97} \\ \hline
	\end{tabular}}
	\label{Tab:Table7}
\end{table*}

\begin{table*}[htp!]
	\caption{Analysis of different pre-trained models on RSSCN7 $\rightarrow$   UC Merced}
	\centering
	\resizebox{1.0\textwidth}{!}
	{
	\begin{tabular}{ccccccccc}
		\hline
		\multirow{2}{*}{Methods}                                & \multicolumn{8}{c}{RSSCN7 $\rightarrow$   UC Merced}                                                                                     \\ \cline{2-9} 
		& Farmland       & Forests        & Dense residential & Rivers         & Parking        & Unknown        & Avg\_Shared    & Avg            \\ \hline
		SDG-MA (Pre-trained model on   ImageNet)              & \textbf{92.00} & 37.00          & 51.00             & 69.00          & 79.00          & \textbf{21.56} & 65.60          & 58.26          \\
        SDG-MA (Pre-trained model on RSSCN7 without initial weights trained on ImageNet) & 46.00          & 23.00          & 6.00              & 2.00           & 13.00          & 7.63           & 18.00          & 16.27          \\
		Ours (Pre-trained model on RSSCN7 with initial weights trained on ImageNet) & \textbf{92.00} & \textbf{64.00} & \textbf{63.00}    & \textbf{76.00} & \textbf{93.00} & 17.13          & \textbf{77.60} & \textbf{67.52} \\ \hline
	\end{tabular}}
\label{Tab:Table8}
\end{table*}

\subsubsection{Ablation Study of Model Adaptation}
In order to verify the efficacy of the proposed sample-level transferability weight, we perform  ablation studies that evaluate variants of SDG-MA, which are listed in Tables~\ref{Tab:Table2}, \ref{Tab:Table3}, ~\ref{Tab:Table4}, and~\ref{Tab:Table5}. SDG-MA w/o d is the variant that does not integrate the domain similarity into the sample-level transferability weight. SDG-MA w/o y is the variant that does not integrate confidence into the sample-level transferability criterion. As shown in Tables~\ref{Tab:Table2},~\ref{Tab:Table3},~\ref{Tab:Table4}, and~\ref{Tab:Table5}, SDG-MA outperforms SDG-MA w/o d and SDG-MA w/o y, which indicates that both the domain similarity and the confidence in the transferability weight are necessary and important for UniDA tasks. 

\subsubsection{Decision Threshold Analysis}
The hyperparameter $w_{0}$ is used to decide whether the model would label a sample as ``Unknown" or use the predicted label. We analyze two cases of  $\xi < 0.2$ (RSSCN7 $\rightarrow$ UCM) and $\xi > 0.2$ (AID $\rightarrow$ NWPU), which are described in Fig.~\ref{fig:6}(a) and Fig.~\ref{fig:6}(b), respectively. 

As shown in Fig.~\ref{fig:6}, ``Target domain" represents the average accuracy of all classes, which measures the generation ability of the source domain space and the domain adaptation ability of the model. ``Target-unknown" is the target accuracy of the ``Unknown" class, which is a crucial metric for evaluating the vulnerability and robustness of the model. Note that there are large differences in the results for a threshold in a wide range between 0 and 2.0.
When $\xi$ is less than 0.2 (Fig.~\ref{fig:6}(a)), the average accuracy of the target domain maintains a high and stable accuracy between 0 and 0.8, and target-unknown rises significantly after 0.4. Thus, $w_{0}$ can be set to 0.8 for the case of $\xi < 0.2$. Furthermore, when $\xi$ is greater than 0.2 (Fig.~\ref{fig:6}(b)), the average accuracy of the target domain exhibits  little variance at higher values in a wide range
between 0 and 1. However, the accuracy of target-unknown increases significantly after exceeding 0.8. Thus, in order to ensure a positive comprehensive accuracy when the case of $\xi > 0.2$,  $w_{0}$ can be set to 1.
\subsubsection{Varying Size of Shared Label Sets}
We explore the effect of the percentages of shared and private label sets  on SDG-MA by varying the size of $Y$. This is done on AID $\rightarrow$  NWPU. Fig.~\ref{fig:6}(c) shows the accuracy of SGD-MA with different $Y$. When $Y$ = 0, the source domain and target domain have no overlap on label sets, i.e. $Y_f \cap Y_t=\emptyset$. It is observed that SDG-MA classifies all categories into ``Unknown". Furthermore, when $Y$ keeps increasing, the performance of SDG-MA remains stable and has high precision. It has demonstrated that SDG-MA is robust for different percentages of shared and private label sets.
\subsubsection{Ablation Study of Source Data Generation}
We go deeper into the efficacy of the proposed SDG by performing an ablation study that evaluates the data diversity module. The results  on RSSCN7 $\rightarrow$ UCM and AID$\rightarrow$ NWPU are  shown in Tables~\ref{Tab:Table6} and~\ref{Tab:Table7}, respectively. Our proposed SDG-MA performs better than SDG-MA without classifier loss and SDG-MA without style loss, indicating both the classifier loss  and the  style loss in the data diversity module are crucial  and necessary for synthetic source data generation. More specifically, when the  classifier loss and the style loss are not considered for SDG-MA, both category and overall accuracy are relatively poor. This phenomenon indicates that the restoration of data content and ensuring data diversity are paramount to the generation of reliable data. In addition, SDG-MA without style loss  outperforms SDG-MA without classifier loss, meaning that the classifier loss (recovering the data content) is even more crucial.

\subsubsection{Comparison of Different Data Diversity Generation Schemes}
In the SDG stage, data diversity is the key to the successful generation of source data distributions. Recently, two mainstream methods have been applied to ensure the diversity of data generation.  The first,  GAN-based methods (such as 3C-GAN~\cite{li2020model}), are used to produce target-style training samples. Specifically, a discriminator is introduced to match the
distributions between the target samples and the generated source samples through the use of adversarial training. 
The second, a decoder loss in KEGNET~\cite{yoo2019knowledge}, produces
similar data points for each class and increases the pairwise distance between sampled data points. We compare our proposed style loss in data diversity module with these two methods on RSSCN7 $\rightarrow$ UCM and AID$\rightarrow$NWPU. The results are presented in Tables~\ref{Tab:Table6} and~\ref{Tab:Table7}, respectively. It can be seen that the generating ability of our proposed  style loss is significantly better than that of 3C-GAN~\cite{li2020model} and KEGNET~\cite{yoo2019knowledge}, with respect to solving the problem of source domain generation in UniDA without source data. In addition, our style loss maintains a more prominent and uniform performance in per-class accuracy. It has been demonstrated that data points generated by the style loss have better intra-class and inter-class diversity.
\subsubsection{Ablation study  of the pre-trained model in SGD} An  ablation study  of the pre-trained model is conducted to  investigate the effect of the pre-training models with different datasets and different initializations on the source data generation. The results on RSSCN7 $\rightarrow$   UC Merced are presented in Table~\ref{Tab:Table8}. It is worth noting that for the pre-trained model on ImageNet~\cite{russakovsky2015imagenet}, the feature extractor comes from the pre-trained ResNet-50 on ImageNet, and the classifier is from the pre-trained ResNet-50 on RSSCN7, in order to ensure the condition of the UniDA setting. Compared with our proposed SGD-MA (pre-trained model on RSSCN7 with initial weights trained on ImageNet), 
it is obvious that the overall average of SDG-MA by using the source data generated from the pre-trained model on RSSCN7  is better than that of SDG-MA by using  the pre-trained model on ImageNet. Furthermore,
we can observe that initial weights trained on ImageNet have a relatively large impact on the proposed source domain generation module, by comparing the pre-trained model without initial weights trained on ImageNet and the pre-trained model with initial weights trained on ImageNet. Thus, we can conclude that both the pre-trained model based on ImageNet (natural images)  and the pre-trained model based on RSSCN7 (remote sensing images) have an impact on the proposed source data generation. The pre-trained model based on ImageNet is used to provide reasonable initial weights of the feature extractor, and the pre-trained model based on RSSCN7 provides an effective category distribution of remote sensing images for the proposed SGD.

\section{Conclusions}
We have introduced a novel Universal Domain Adaptation setting for remote sensing image scene classification, including UniDA with source data (MA) and UniDA without source data (SDG-MA). UniDA removes all constraints on the relationship between label sets of the source and target domains, 
which has a high practical value and promotes the development of DA in remote sensing. To realize universal domain adaptation with or without source data, a dual-stage framework is proposed, consisting of a source data generation stage  and the purpose of the  model adaptation stage.
The source data generation stage is to estimate the conditional distribution of the source data and  generate reliable  synthetic source images from both data content and data style, when the source data is not available. Furthermore,  the model adaptation stage aims to detect samples from the target shared label sets and those in target private label sets  utilizing the proposed transferable weight. This work can serve as a starting point in a challenging UniDA setting for remote sensing images. However, it is difficult for the transferable weight in model adaptation to  tune an optimal threshold to apply it to all UniDA tasks of remote sensing images. Thus, in the future, we will focus on adaptively learning the threshold through the use of an open-set classifier.


\bibliographystyle{IEEEtran}
\bibliography{egbib}
\end{document}